\RequirePackage{fix-cm}
\documentclass[smallextended]{svjour3}       
\smartqed  
\usepackage{graphicx}
\usepackage[numbers,sort&compress]{natbib}
\usepackage{multirow}
\usepackage{booktabs}
\usepackage[table,xcdraw]{xcolor}
\usepackage{subfig}
\usepackage{amsmath}
\usepackage{hyperref}
\usepackage{enumitem}
\usepackage{xcolor}
\usepackage{footnote}
\usepackage{tablefootnote}
\newcolumntype{M}[1]{>{\centering\arraybackslash}m{#1}}

\makesavenoteenv{table} 
\begin{document}

\title{Connectionist Recommendation in the Wild: On the utility and scrutability of neural networks for personalized course guidance}
\titlerunning{Connectionist Recommendation in the Wild}        

\thanks{This work was supported by grants from the National Science Foundation (Awards 1547055 \& 1446641)}

\author{Pardos\and
        Fan\and
        Jiang
}


\institute{Zachary A. Pardos \at
              University of California, Berkeley\\
              Berkeley, CA 94720\\
              \email{pardos@berkeley.edu}  
              \and
			  Zihao Fan\at
              University of California, Berkeley\\
              \email{zihao\_fan@berkeley.edu}
              \and
              Weijie Jiang \at
              University of California, Berkeley\\
              \email{jiangwj@berkeley.edu}
}

\date{Received: date / Accepted: date}

\maketitle

\begin{abstract}
The aggregate behaviors of users can collectively encode deep semantic information about the objects with which they interact. In this paper, we demonstrate novel ways in which the synthesis of these data can illuminate the terrain of users' environment and support them in their decision making and wayfinding. A novel application of Recurrent Neural Networks and skip-gram models, approaches popularized by their application to modeling language, are brought to bear on student university enrollment sequences to create vector representations of  courses and map out traversals across them. We present demonstrations of how scrutability from these neural networks can be gained and how the combination of these techniques can be seen as an evolution of content tagging and a means for a recommender to balance user preferences inferred from data with those explicitly specified. From validation of the models to the development of a UI, we discuss additional requisite functionality informed by the results of a usability study leading to the ultimate deployment of the system at a university.   
\keywords{Recommender systems \and Distributed representation \and Recurrent neural networks \and Skip-gram \and Scrutability \and Usability study \and Higher education}
\end{abstract}

\section{Introduction}
\label{intro}
Adaptive recommendation systems \cite{brusilovsky1996methods} have utilized user ratings as a collaborative source of user preference as well as item content and tags to make semantic based recommendations. Recent connectionist (i.e., neural network) approaches to representing items are revealing that semantics can be learned implicitly from behaviors \cite{pardos2018map,chen2018behavior2vec}. The most salient example of this has been its application to language, where words are embedded into a vector space based on the collection of word contexts observed in a large text corpus \cite{mikolov2013distributed}. While words themselves have lexical semantics, the space they are embedded into consists of conceptual semantics \cite{hinton1986learning} such that distributed representations of royalty, capital of country, and many other relationships can be found as features of the space. Applications outside of language have shown that there is a semantic coherence to embeddings formed from sequences of tutor problem solving \cite{pardos2017imputing,Pardos2018}, movie viewing \cite{barkan2016item2vec}, e-commerce clickstream \cite{chen2018behavior2vec}, and course enrollments \cite{pardos2017school,pardos2018map}. The significance to the user modeling community is that these representations of items, learned implicitly from behaviors, can serve as an alternative, and at times more detailed source of semantics of an item that can be inferred in place of explicit tagging, which can be expensive or otherwise untenable. 

Methodologically, we demonstrate an application of the inferences of these implicit tags being used in place of conventional content-based recommendation. Furthermore, we show how this representation of items, akin to a knowledge-base, can be combined in a hybrid fashion \cite{burke2002hybrid} with recurrent neural networks (RNN) to construct a recommendation system that attempts to balance this new content-based recommendation with a more traditionally collaborative one. Evaluation of the RNN was designed to mimic the real-world scenario where courses are being suggested for a student's next semester. We discuss and methodologically address issues of scrutability in both model types. While both models utilized are neural network architectures, word embedding models are linear and therefore create a vector space with arithmetic and scalar closure which can be queried and reasoned about. Recurrent neural networks, and other ``deep" \cite{lecun2015deep} models, contain a high degree of non-linear transformations, such as the standard sigmoid activation or the rectified linear activation used in image classification \cite{krizhevsky2012imagenet}. This high degree of non-linearity makes them naturally more opaque \cite{burrell2016machine}; however, progress has been made towards visualizing the manifold space of neural networks, and we demonstrate how these techniques can be used to extract substantive information from our RNN model. 

In post-secondary education, particularly four-year degree programs, it is important for institutions to strike a balance between guidance and information; providing enough guidance to minimize poor enrollment decisions while allowing students enough choice to develop self-regulation and find their own path given the information at their disposal. In our system implementation, we use an RNN with course requirement filter options to provide guidance and a skip-gram model to provide course relevancy information with respect to their stated interests. The current landscape of enrollment systems and course catalogues, despite being at the core of students' interface to mapping their educational paths, remains a web 1.0 experience for most with little to no historical data or even classical recommendation models used to enhance users' experiences. We contribute a generational leap in this application area, developing a personalized course guidance and information system bringing to bear the aforementioned nascent modeling paradigms. We conduct a usability study to understand the enrollment priorities and existing sources of course information of students and to better align the system features and interface with their needs. The system serves as not only a proof-of-concept of the this hybrid representational approach to recommendation, but as a fully deployed system, accessible by every major on campus and with real-time connection to enrollment and class schedule APIs. 

We conclude the paper with a discussion of the limitations of this modeling approach, the potential for persisting undesirable historic behaviors, and strategies for counteracting these undesirable biases.

\section{Related Work}
In this section, we overview the relevant literature on student achievement in the US post-secondary context, the collaborative and content-based algorithms underpinning most recommender systems outside of education, and the existing recommendation methods and systems as they have been applied to the post-secondary context. The related literature on scrutability will be introduced in section 7.1, after presentation of results of a user study. 
\label{sec:1}

\subsection{The Need for Scalable Guidance in Higher-Education}




The normative time for earning a Bachelor’s degree in the United States has wandered far from the expected four-year target. \citet{deangelo2011completing} found that 6 years after matriculation, only 49.5\% of students at public colleges had earned their degree compared with 78.2\% at private universities. Extracurricular time commitments play a large role, with full-time students far more likely to complete their degree than part-time students who often hold a job while pursuing their degree \cite{shapiro2017completing}. However, part of the problem of student post-secondary degree attainment is attributable to matters of guidance \cite{cca}, with a national adviser to student ratio of one to 400 and the observation that a semester's worth of non-requirement fulfilling credits are taken by students. There is evidence that more readily available quality guidance on course selection could most benefit post-secondary students in two year degree programs \cite{hodara2016improving} whom the United States Government Accountability Office (GAO) has declared\footnote{\url{http://www.gao.gov/assets/690/686530.pdf}} are in need of greater access to information as they navigate their way to transferring to a Bachelor's program.

\label{sec:2}

\subsection{Personalization \& Recommendation}
Definitions of personalization, adaptation, and individualization and their relationships to one another are a topic of frequent discussion in the field \cite{aleven2016instruction}. Adaptation can be seen as the ability of a system to change based on the actions of its users. This change could be one that affects all users of the system; for example, changing the sequence of an online course syllabus (for everyone) based on an analysis of learner behaviors from the previous offering of the course. If the sequence of the course were changed, catering to learners at the individual level, this would be an example of adaptation and personalization \cite{pardos2017enabling}. While adaptation necessarily involves changes based on data, personalization does not have this as a requirement, as decisions about how to personalize can come from the system, the user, or both. The term ``individualization", on the other hand, refers to adaptation specifically to one's prior domain knowledge and is applied in Intelligent Tutoring Systems and other instructional approaches based on skill mastery \cite{corbett2001cognitive}. Adaptive personalization can be used to determine a student's path or available options; alternatively, when implemented in the form of a recommendation, it takes on a more informative objective, allowing the user to make the final decision and not restricting the options available to them. The appropriateness of these restrictions is dependent on context, as learners in secondary school may have less developed self-regulation \cite{zimmerman1990self} or executive function \cite{diamond2011interventions} and require a greater degree of structure than in post-secondary. At either educational level, the importance of student choice is a topic of active discussion, particularly in its relation to the development of agency \cite{snow2015does}. 

\subsubsection{Collaborative Filtering} 

Collaborative Filtering is the process of filtering or evaluating items using the opinions
of the crowd \cite{schafer2007collaborative}. It often involves factorizing a user and an item based on historic ratings and then imputing users' ratings of items based on similar users in the factor space. The general approach has come out of e-commerce \cite{linden2003amazon} with the formalization of matrix factorization, now synonymous with collaborative filtering, introduced in a movie prediction competition \cite{koren2009matrix}. These works catalyzed adoption in adaptive systems in the user modeling community \cite{guo2012simple, berkovsky2007cross, carmagnola2009sonars, konstan2012recommender} and beyond \cite{paterek2007improving, mairal2010online, koren2010factor, koren2010collaborative, koren2015advances}.
The data presented to these models involved explicit feedbacks (ratings). However, much of users' interactions with items do not involve ratings and the amount and availability of ratings can be quite scarce. For these reasons, recent attention is increasingly shifting towards implicit data \cite{he2016fast, liang2016modeling, bayer2017generic, he2017neural}. However, implicit data is more challenging to utilize for matrix factorization due to the natural scarcity of negative feedback \cite{he2016fast}. 

Neural networks have, in contrast, performed well in predicting sequences of implicit or behavioral data, beginning most notably with language \cite{mikolov2010recurrent} and extending to machine translation \cite{sutskever2014sequence}, text classification \cite{lai2015recurrent}, speech recognition \cite{graves2013speech}, speech synthesis \cite{fan2014tts}, image captioning \cite{mao2014deep}, image generation \cite{gregor2015draw}, learner sequence prediction \cite{tang2016modelling}, and more. They are beginning to be used as a form of collaborative filtering for within-session recommendation in e-commerce \cite{jannach2017session,hidasi2015session, tan2016improved} and across-session recommendation based on longer histories of user interaction \cite{devooght2017long}. 

\subsubsection{Content-based Recommendation}

Content-based recommendation approaches recommend items with similar attributes to the items that a user has liked in the past. They have been used in a variety of domains ranging from recommending web pages \cite{van2000using}, to restaurants \cite{pazzani1999framework}, news articles \cite{phelan2009using, abel2011analyzing}, and television programs \cite{cragun1999dynamic, yu2006tv}. Many of these approaches are based on user and item representation. In \citet{abel2011analyzing}, a Twitter user profile and various tweets are used to enrich the semantics of individual Twitter activities and allows for the construction of different types of semantic user profiles, which are then applied to content-based recommendation for recommending news articles. Recently, connectionist approaches (i.e. neural networks), such as RNNs, have also been utilized for content-based recommendation. A deep architecture adopting Long Short Term Memory (LSTM) networks, a variant of RNNs, was used to jointly learn two embeddings representing the items to be recommended as well as the preferences of the user \cite{suglia2017deep}. The details of content-based recommendation systems differ based on the representation of item and user profiles. Item representations include (1) structured data (e.g. a database of item properties) (2) unstructured item descriptions and (3) word embeddings \cite{musto2016learning}. User profiles have included (1) a model of the user’s preferences and (2) a history of the user’s interactions with the recommendation system. This may include storing the items that a user has viewed together with other information about the user’s interaction. Information can also be explicitly provided in content-based recommendation systems via user customization, by providing an interface that allows users to construct a representation of their own interests \cite{pazzani2007content}. 

\subsubsection{Distributed Semantic Representation}
Item semantics have come from user tags and the content of the item itself. Representation learning is demonstrating an ability to learn aspects of the semantics of items based on behavioral data. This demonstration began with the application of skip-grams and continuous-bag-of-words models to language \cite{mikolov2013efficient}, where 61\% of pre-defined syntactic and semantic relationships between words were captured by the model trained on a corpora of 1 billion words from Google News articles. These models have also obtained state-of-the-art results on a plethora of Natural Language Processing tasks \cite{levy2014neural, levy2014dependency, goldberg2016primer}. Words close together in the vector space formed by the model can be considered to be synonymous and share a similar semantics and location in the space due to the similarity of their contextual usages. This idea can be ported to recommendation of products, where products close together in the vector space can be assumed to share features with one another \cite{grbovic2015commerce}. An extension of this leverages existing item metadata to regularize the item embeddings, which  outperforms previous approaches to recommendation \cite{zanotti2016infusing}. Item and user representations can also be learned via representation learning models such that any timely ordered sequence of items selected by a user will be represented as a trajectory of the user in a representation space \cite{guardia2015latent,luo2018diag}. The distinction between this representation of a user and those derived from a matrix factorization approach is that these form a vector space which can be arithmetically manipulated while retaining its semantic properties, giving it great flexibility in allowing users to express complex relational preferences. These representations also differ from matrix factorization in their type of source data, using behavioral (implicit) data instead of ratings (explicit).

\subsubsection{Recommendation in Higher Education}
Several paradigms of approaches have been taken towards improving the course enrollment experience for students. \citet{parameswaran2011recommendation} built a recommender based on constraint satisfaction; taking into account institutional breadth and degree requirements as well as scheduling constraints of the student and courses being recommended. \citet{li2012visual} proposed a course to proficiency tagging regime combined with a visual mapping between proficiency components for guiding students through a pathway of post-requisites. \citet{farzan2011encouraging} asked students to give their career goals and then rate courses for their workload and relevance to those goals, allowing other students to then select courses based on those characteristics.

A related body of work has focused on predicting students’ outcome on courses or in programs; using C4.5 decisions trees to help students select courses they are likely to succeed in within the School of Systems Engineering at the University of Lima, Peru \cite{sacin2009recommendation} with a similar feature having existed in a system deployed widely at Stanford, presenting students with grade distributions, popular course sequences, and course evaluation summaries \cite{chaturapruek2018data}. Another approach showed the likelihood of passing to students while taking a course using red, yellow, and green indicators to try to improve engagement at the University of Purdue \cite{arnold2012course}. \citet{jayaprakash2014early} designed an early warning system which predicted if a student was likely to drop-out and deployed social interventions in an effort to target resources towards supporting students most in need. \citet{elbadrawy2016domain} investigated how  student and course features influence enrollment patterns and used these features to define student and course groups at various levels of granularity. \citet{luo2018diag} analyzed student proficiencies and predicted on-time graduation of Molecular and Cellular Biology majors using vector representations learned from the same enrollment histories used in this paper. Data from Massive Open Online courses (MOOCs) have also spurred analysis of learner outcomes in higher-ed, with \citet{whitehill2017delving} using a logistic model on hand engineered features from learner event log data and \citet{yang2013turn} predicted MOOC learner drop-out based on their social positioning as inferred from discussion board activity.

Our methodological approach contributes to this literature by evaluating the suitability of neural networks for next semester course recommendation and for inferring course similarity. The developed system's core features are driven by user adaptive modeling approaches. While other systems detailed in this section employing grade prediction can be considered to be in this same category, our scaling of this type of system across a large institution, serving all of its students and majors, is novel. In the wider scope of recommender systems, neural network approaches have been evaluated methodologically; however, our work detailing the utilization and user testing of these nascent connectionist techniques in a production system is among the first in any domain.

\section{Data Set}
\begin{figure}
\centering
\includegraphics[height=2in, width=3.5in]{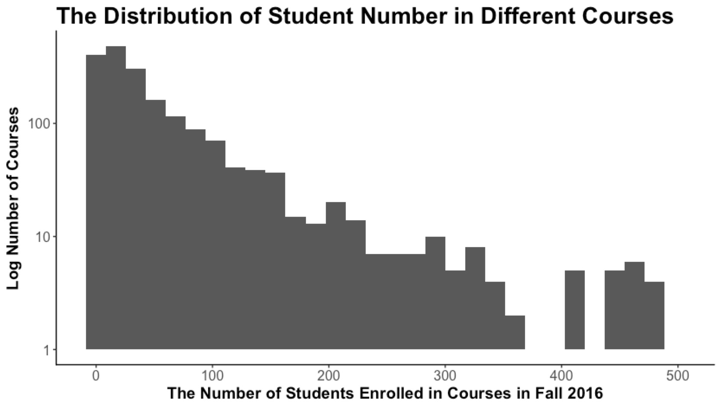}
\caption{The log scale histogram of total enrollments for each class.}
\label{fig1b}
\end{figure}

\begin{figure}
\centering
\includegraphics[height=2in, width=3.5in]{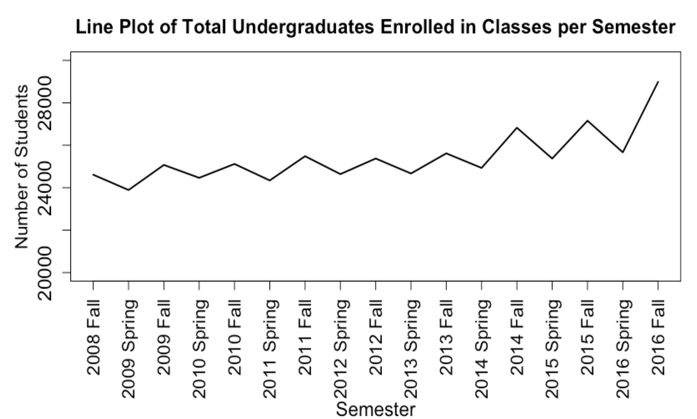}
\caption{Total number of active students by semester.}
\label{fig1}
\end{figure}

We used a dataset from the University of California at Berkeley (UCB) which contained anonymized student course enrollments from Fall 2008 through Fall 2016. The dataset consisted of per-semester course enrollment information for 108,033 undergraduates with a total of 2.2M course enrollment records and 265 different majors. A course enrollment meant that the student was still enrolled in the course at the conclusion of the semester. The median course load during students'’ active semesters was four. There were 9,714 unique lecture courses from 197 subjects in 124 different departments hosted in 17 different divisions of 6 colleges. Course meta-information contained course subject, department name, total enrollment, and max capacity. In all analyses in this paper, we only considered courses with at least 10 enrollments total over the 8 year period. A log scale histogram of total enrollments for each class and the total number of active students by semester are shown in Fig. \ref{fig1b} and Fig. \ref{fig1}, respectively. The raw data were provided in CSV format by the UCB Educational Data Warehouse team. Each row of the course enrollment data contained date stamp information, an anonymous student ID, entry type (transfer student or new freshman), and declared major(s) at each semester. Course information included course name, subject, department, enrollment count, and capacity. The basic structure of the enrollment data is shown in Table \ref{data}, where, for example, a transfer student with anon ID x282243 enrolled in a course called Math 121 in Fall 2014 and received an A.  

\begin{table}[]
\centering
\caption{Sample of data from the dataset of student enrollments.}
\label{data}
\begin{tabular}{M{4em}|M{5em}|M{4em}|M{5em}|M{4em}|M{5em}|M{3em}}
\hline
\rowcolor[HTML]{333333} 
{\color[HTML]{FFFFFF} \textbf{Semester Year}} & {\color[HTML]{FFFFFF} \textbf{STU ID (anon)}} &
{\color[HTML]{FFFFFF} \textbf{Major}} &
{\color[HTML]{FFFFFF} \textbf{Entry Type}} & {\color[HTML]{FFFFFF} \textbf{Subj.}} & {\color[HTML]{FFFFFF} \textbf{Course Number}} & {\color[HTML]{FFFFFF} \textbf{Grade}}  \\ \hline
\rowcolor [HTML]{C0C0C0}
Spring 2014 & x137905  & Law & New Freshman & Law & 178 & B  \\ \hline
Summer 2014 & x137905  & Law & New Freshman & Law & 165 & C  \\ \hline
\rowcolor [HTML]{C0C0C0}
Fall 2014 & x282243 & Math & Transfer Student & Math & 140 & B+  \\ \hline
Fall 2014 & x282243 & Math & Transfer Student & Math & 121 & A  \\ \hline
\end{tabular}
\end{table}

\section{Vector Space Representation of Courses}
In this section, we describe the skip-gram model and how we applied it to our context of course enrollments. We validate the model against a list of course equivalencies kept by the registrar to investigate if courses similar to one another in the vector space are also similar according the registrar specifications. These equivalencies describe courses that were determined to be too similar to one another in the material they teach for students to be able to earn full credit for both. 
\subsection{Formal Course2Vec Definition Using Skip-grams}
\begin{figure}
\centering
\includegraphics[height=3in, width=3.5in]{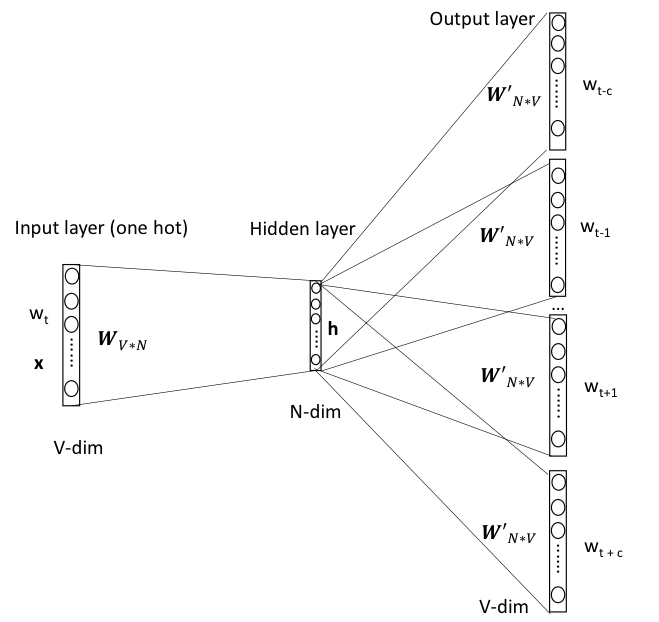}
\caption{Skip-gram model architecture.}
\label{skip-gram}
\end{figure}

We use a skip-gram model (Fig. \ref{skip-gram}) to generate course representations from student enrollment sequences. In the same way that the meaning of a word can be inferred from its usage, we infer the meaning (or representation) of a course as a function of its context. For every student, $s$, their chronological course sequence, $T$, is produced by first sorting by semester then randomly shuffling their within-semester course order. Negative affects of this randomization are diminished, as the order of course tokens is not taken into account in the output context window of the model, described below.

In a skip-gram, the vector representation of an input course is defined as:
\begin{equation*}
\boldsymbol{v}_{w_I} = \boldsymbol{W}^T\boldsymbol{\delta}(w_I)
\end{equation*}
Where $\boldsymbol{W}^T$ is the left side weight matrix in Fig. \ref{skip-gram}, indexed by the one-hot of the input course, $\boldsymbol{\delta}(w_I)$. A softmax layer, typical in classification tasks, is used to produce a probability distribution over courses to predict courses in context:
\begin{equation*}
p(w_o|w_I) = \frac{{\rm exp}(\boldsymbol{W}'\boldsymbol{\delta}(w_o)\boldsymbol{v}_{w_I})}{\sum_{j=1}^V {\rm exp}(\boldsymbol{W}'\boldsymbol{\delta}(w_j)\boldsymbol{v}_{w_I})}
\end{equation*}

For a given output course, $w_o$, in the vocabulary, its probability is the exponential normalization defined by the exponentiation of the input course's vector, $\boldsymbol{v}_{w_I}$, multiplied by the output course's vector, $\boldsymbol{W}'\boldsymbol{\delta}(w_o)$, divided by the sum of all courses' exponentiation of their output vector multiplied by the input vector. An output vector is the multiplication of the right side weight matrix, $\boldsymbol{W}'$, with a one-hot of the output course, $w_o$.

The model weights are fit to the data using stochastic gradient descent optimizing a cross-entropy loss objective function across all students' sequence of courses:

\begin{equation*}
C = -\sum_{s\in S} \frac{1}{T} \sum_{t=1}^T \sum_{-c\le i \le c, i\ne 0} {\rm log} p(w_{t+i}|w_t)
\end{equation*}

Where, for each student, $s$, the average loss is calculated over the input courses at each position, $t$. The loss for a single input course of a student is the sum of the log of the model'’s probability of observing the courses within a time slice window size $c$ positions to the left and right of the current position, $i$, in the student's serialized sequence of courses, $T$. A value of $c$ of 2, for example, would span two courses before and after the input course, capturing the median number of courses taken within a semester. When $t = 1$ (the beginning of the sequence), the context window only involves $t+1$ through $t+c$. Similarly, when $t = |T|$ (the end of the sequence) the context window only involves $t-c$ through $t-1$. While the model is trained to minimize error in predicting the courses in context, the intended extract from the model after training is not its predictions, but rather the learned representations of the courses in the form of the weight vectors associated with each course. These weight vectors, which also comprise the hidden layer activations of the model for each course, are the automatic featurization of the course. Courses which have similar contexts become mapped to vectors of similar magnitude and direction in order to minimize the loss. 

\subsection{Equivalency Validation Result}
\begin{figure}
\centering
\includegraphics[width=\columnwidth]{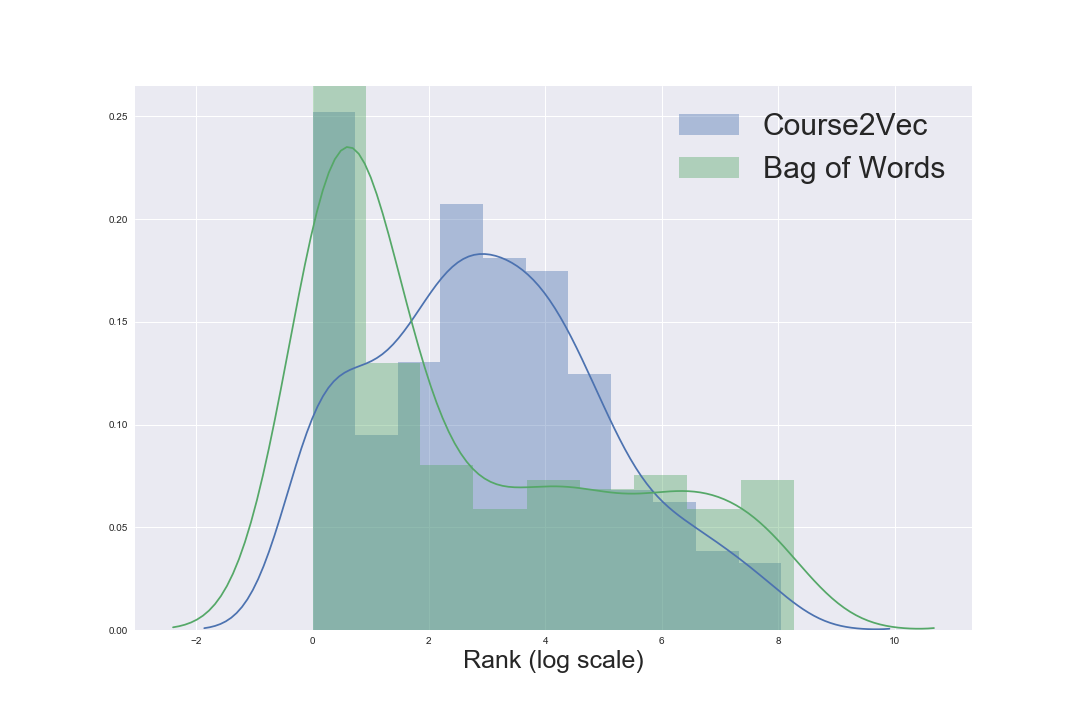}
\caption{Validation set distribution of ranks (in log scale) for Course2Vec and Bag of Words models. Lower ranks suggest closer adherence of the course representations to the ground truth similarity information provided by the Registrar.}
\label{course2vecdist}
\end{figure}

If the skip-gram model successfully captured concepts in the representation of courses, then courses teaching similar material should be similar to one another in the space. We use a skip-gram whose hyperparameters were optimized to this similarity metric in prior work \cite{pardos2018map}. As a comparison, we also construct a bag-of-words (BOW) vector for each course based on its course description. Ideally, courses teaching similar material should also be close to one another in their course descriptions; however, descriptions are sometimes outdated or very brief and therefore may not reflect the true overlap of material. Pre-processing of the BOW approach included word stemming and punctuation and stop word removal. The most frequent 15 words were also removed as they were highly generic, consisting of words such as ``students" and ``course."
The validation set we used for this analysis was a list of pairs of courses that were considered as credit equivalent by faculty, and enforced by the UC Berkeley Registrar. We filtered out pairs between cross-listed courses where the course descriptions would be exactly the same. After filtering, 461 equivalency pairs remained for validation. The representation, which we coin "Course2Vec," was a 229-dimension vector while the bag of words representation was a 9,102-dimension multi-hot vector for each course. We ran our experiments with both Course2Vec representations and the bag of words representation on this validation set and used cosine similarity as our similarity metric. For each pair of equivalent courses $(c1,\ c2)$, we denote the representation of the courses as $(v_1,\ v_2)$ and we calculate the rank of $cosine(v_1, v_2)$ in the list of $[cosine(v_1,\ v_i)]$, where $i$ goes through the 3,939 courses that had both Course2Vec and BOW representations. A rank of 1 meant that the predicted most similar course to the input course from the validation set was in fact the course listed as equivalent to it by the Registrar. The statistics of the results are listed in Table \ref{equivalencytest} and the distribution of ranks is shown in Figure \ref{course2vecdist}. We can tell from the statistics that both methods performed well. Bag-of-words has a better median rank but the distribution of the Course2Vec model shows that it performs more consistently on all the courses while performance of the bag-of-words model is more prone to performance at the extremes. Interestingly, this signals that there is similarity information conveyed strictly from course enrollment behaviors that is not encoded in course semantic descriptions. 

\begin{table}[htbp]
  \centering
  \caption{Equivaliency validation results of the two course representations.}
    \begin{tabular}{l|l|l|l}
  
    \textbf{Course Representation} & \textbf{Median Rank} & \textbf{Average Rank} & \textbf{Std of Rank} \\
    \hline 
    Course2Vec  & 18 & 148 & 413 \\
    Bag of Words & 4   & 291 & 744 \\
 
    \end{tabular}%
  \label{equivalencytest}%
\end{table}%

\section{Collaborative-based Recommendation Using RNNs}
In this section, we apply Recurrent Neural Networks (RNNs) to the task of recommendation. In this context, RNNs can be seen as a type of collaborative filtering method with a non-linear embedding serving as the factorization of items and the cumulative hidden state, a non-linear function of the current and previous inputs, a factorization of the user over time.

The predictions made by our RNN models provide peer-like suggestions representing the common next enrollments by students like them, as learned from historic enrollments. Our prediction task is similar to that of text generation in computational linguistics; however, instead of using words or characters as the input and output, we use courses. In our models, each semester is considered a time slice, as there is no discernible order of courses within a semester in our data, and thus we use a multi-hot representation to express multiple courses occurring within a semester. 

In order to be included in training, we imposed that the student had to have at least two semesters of data in the training set (Fall 2008 through Summer 2016).  We also ruled out all students with more than 12 semesters (students who attended more than 4 years, including all 4 summer sessions). This filtering allowed the recommendation to be slightly biased towards the enrollment behaviors of students who graduated approximately on time.

RNNs have been a successful modeling paradigm in a variety of recommendation tasks and have several properties appropriate to our context. Firstly, information about the temporality of course enrollments is assumed to be useful in predicting course selections. That is, we assume that there are cases in which students who enrolled in a course (e.g. Linear Algebra) the first semester may have a different distribution of courses taken in the fifth semester than students who took that same course in the fourth semester. RNNs allow for modeling of temporality with their recurrence relation, which applies a constant transformation at every timeslice. Many courses are enrolled in by fewer than 100 students and subsequently, very few students share the exact same course selection history with each other. Therefore, a second desirable property of a model applied to our context is to be able to factorize courses such that they can be related to one another and, in turn, related students to one another. RNNs, and all other connectionist topologies, provide effective generalization of items through an embedding.

We used three modeling techniques in our experiments; RNNs, N-gram models, and two  models based on course popularity as baselines.  


\subsection{Recurrent Neural Network Architecture and Optimization}
We use a popular variant of RNNs called Long Short-Term Memory (LSTM) \cite{hochreiter1997long}, which helps RNNs learn temporal dependencies with the addition of several gates which can retain and forget select information. They have been shown to generalize using long sequences more effectively than RNN networks \cite{bengio2003neural, gers1999learning} by using a gated structure to mitigate the vanishing gradient phenomenon. While our sequences are not long, our prediction task can benefit from the ability to selectively treat certain course selections as irrelevant (i.e. forgettable) to the prediction of future course selection. Regular RNNs do not possess the ability to ignore certain course selections based on context and in early prototyping, we found them to have equal or lower performance than LSTMs. The decision of what information to forget at time step t is made by the forget gate $f_t$, where $\boldsymbol{w}_t$ is the course representation at time step t and $\boldsymbol{h}_{t-1}$ is the output of the network at time step t-1.
\begin{equation*}
f_t = \sigma (\boldsymbol{W}_{fw}\boldsymbol{w}_t + \boldsymbol{W}_{fh}\boldsymbol{h}_{t-1} + \boldsymbol{b}_f)
\end{equation*}
The decision of which information will be stored into the cell is determined by calculating the input gate $i_t$ and a candidate value $\widetilde{\boldsymbol{C}_t}$ to be added to the state.
\begin{equation*}
i_t = \sigma(\boldsymbol{W}_{iw}\boldsymbol{w}_t + \boldsymbol{W}_{ih}\boldsymbol{h}_{t-1} + \boldsymbol{b}_i)
\end{equation*}
\begin{equation*}
\widetilde{\boldsymbol{C}_t} = {\rm tanh}(\boldsymbol{W}_{Cw}\boldsymbol{w}_t + \boldsymbol{W}_{Ch}\boldsymbol{h}_{t-1} + \boldsymbol{b}_C)
\end{equation*}
We denote $\boldsymbol{C}_t$ as the internal cell state and update it by the intermediate results calculated below. 
\begin{equation*}
\boldsymbol{C}_t = f_t \times \boldsymbol{C}_{t-1} + i_t \times \widetilde{\boldsymbol{C}_t}
\end{equation*}
Finally, we calculate the output of time step t, denoted as $\boldsymbol{h}_t$.
\begin{equation*}
o_t = \sigma(\boldsymbol{W}_{ow}\boldsymbol{w}_t + \boldsymbol{W}_{oh}\boldsymbol{h}_{t-1} + \boldsymbol{b}_o)
\end{equation*}
\begin{equation*}
\boldsymbol{h}_t = o_t \times {\rm tanh}(\boldsymbol{C}_t)
\end{equation*}

The LSTM next course prediction models used in this paper were implemented using Keras \cite{chollet2015keras}, with Theano backend \cite{bastien2012theano, bergstra2010theano}. We implemented a  LSTM course prediction model using a multi-hot representation of course enrollments per timeslice, then added additional input features including major, GPA, and entry type to further enhance its predictions. 

For the simplest course prediction model, we considered different semesters as different time steps. Our model takes all the courses within a semester as input, each course being represented by an index number and therefore corresponding to a position in the one-hot input vector representation. Our input layer directly takes in the summation of all the one-hot vectors of courses, which we call a multi-hot representation, reflecting the multiple courses present in a single semester. The LSTM layer takes in the multi-hot course representation at each time step and generates a hidden state of student course enrollment histories up to that point. Finally, we use a fully-connected layer with softmax activation to convert this representation into the next semester course probability distribution. 
\begin{equation*}
\boldsymbol{z}_t = \boldsymbol{W}_h\boldsymbol{h}_t + \boldsymbol{b}_{fc} 
\end{equation*}
\begin{equation*}
y^i = \frac{e^{z_t^i}}{\sum_{j\in Courses}e^{z_t^j}}
\end{equation*}

At each time step we feed in the multi-hot representation to the neural network and use the next semester's multi-hot representation as the output labels to calculate the loss to be backpropagated. We use categorical cross-entropy as our objective function. By choosing cross-entropy, we are optimizing the similarity between the distributions of the output of softmax layer and the label:
\begin{equation*}
C = - \sum_{j\in Course} \boldsymbol{t}^j {\rm log}\boldsymbol{y}^j
\end{equation*}

\begin{figure}
\centering
\includegraphics[height=2.7in, width=4.7in]{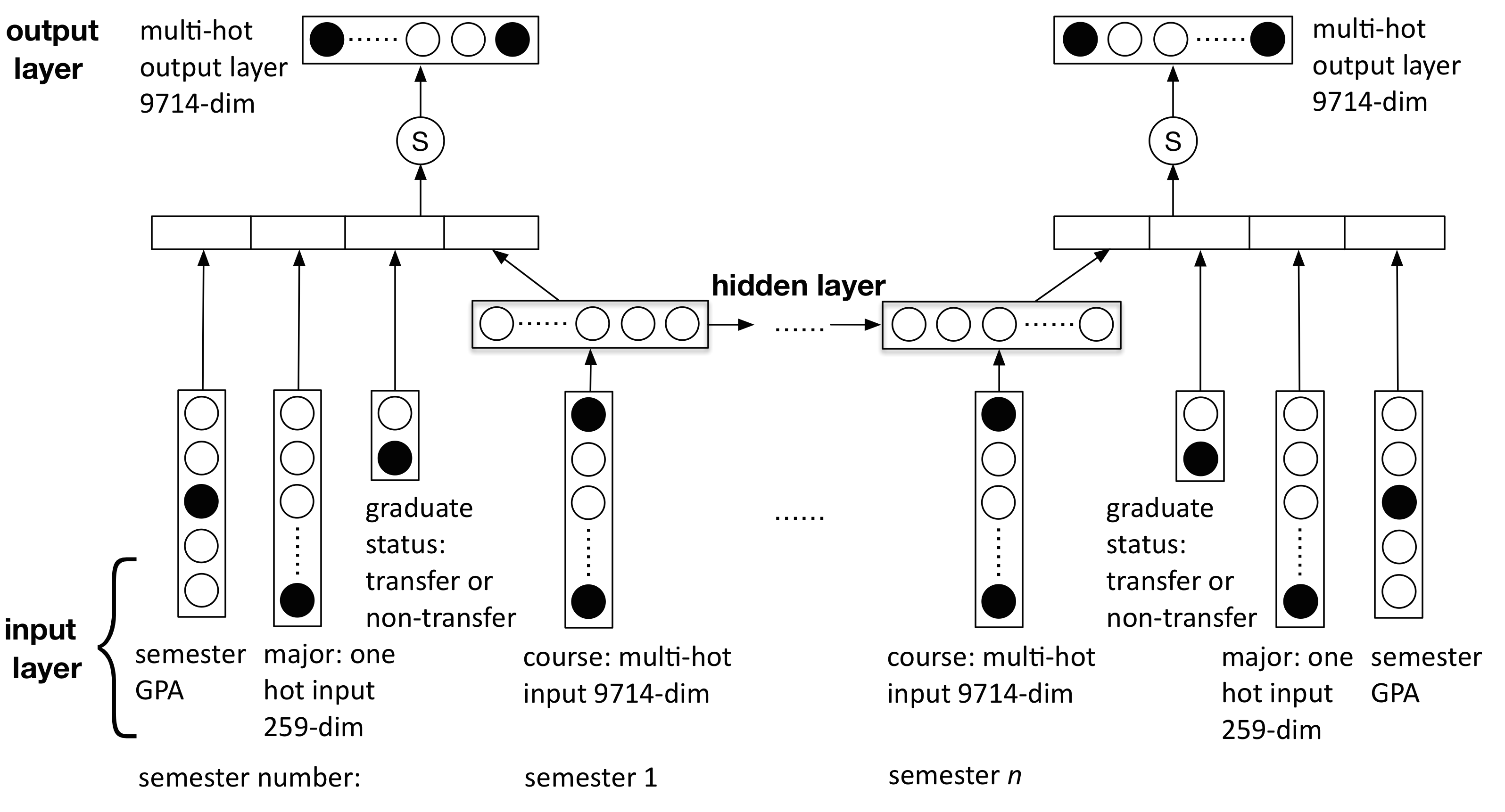}
\caption{The topology of our LSTM course prediction model.}
\label{rnn1}
\end{figure}

To optimize our objective function, we used the Adam optimizer \cite{kingma2014adam} with learning rate of 0.001, $\beta_1= 0.9$, $\beta_2 = 0.999$ and a gradient norm value of 5.0.
One improvement to our simplest model was the incorporation of additional features, including major, entry type, and the last semester's GPA. To achieve this, we needed to modify the RNN hidden state to be concatenated with the one-hot representations of the features, all leading into the fully connected output layer:
\begin{equation*}
\boldsymbol{z}_t = \boldsymbol{W}_h\boldsymbol{h}_t + \boldsymbol{W}_m\boldsymbol{m}_t + \boldsymbol{W}_s\boldsymbol{s}_t + \boldsymbol{W}_g\boldsymbol{g}_t + \boldsymbol{b}_{fc}
\end{equation*}
The structure of the LSTM course prediction model is shown in Fig. \ref{rnn1}.



\subsection{N-gram Model and Popularity Baselines}
N-gram models \cite{brown1992class} are simple yet powerful frequentist models when applied to language modeling tasks. They break sequences from the training set into small sub-sequences with length N as grams. For a given size n, it can approximate the probability distribution of $P(x_n|x_{n-1}x_{n-2}...x_1)$ by building lookup tables for each sub-sequence of length n-1 and their following elements. The number of parameters of N-gram models grows exponentially with the number n. We expect these kinds of models can be competitive course prediction models to our LSTM.


Recommending items based on popularity is a common method used by many prominent online retailers and a common choice for baseline performance in the collaborative filtering literature \cite{rashid2002getting}. We apply two kinds of popularity recommendations; the first is historic popularity of courses in a given semester. For the Fall 2016 semester, for example, courses will be predicted based on the enrollment sizes of all courses in the Fall semesters for which we have data. The second version of popularity baseline we used was to add personalization by major, where courses are sorted by the most historically popular for a given semester among the student's major. Given there are 265 majors, this is a non-trivial level of personalization. Comparison to these popularity baselines can give a sense for the degree of variation that exists in student course selection in general and within a major.

\subsection{Model Evaluation}
Evaluation of the RNN was designed to mimic the real-world scenario where courses are being suggested for a student's next semester. The training set for this evaluation therefore consisted of all course enrollment data prior to and including Summer 2016. The Fall 2016 semester served as the test set, where every student who had course enrollments in the test semester had their courses predicted. This included both students for whom this was their first semester (freshman and new transfers) and students who were continuing from a previous semester of matriculation. For continuing students, their course history was used to make personalized predictions for the test semester, while new freshman, who were all marked as having an 'undeclared' major, would have the same courses predicted since we had no other prior distinguishing information about them in this dataset. New transfer students would also have no prior enrollment data available to us, but they would have a major declared.
In order to perform a hyperparameter search of the RNN without overfitting to the test set, we split the training set into two parts; a validation set consisting of the Fall 2015 semester's enrollments and a sub-training set consisting of enrollments prior to that. We then performed a grid search on the model's hyperparameters and input features on the sub-training set and predicted the validation set. The hyperparameters and feature sets evaluated are shown in Table \ref{parameters}. The median number of courses a student enrolled in for the Fall 2015 semester was four with a high of nine. This meant that the top 10 predicted courses could, in the ideal case, capture all courses students enrolled in. We limited the predictions to only courses offered in the target semester. This was with accomplished by predicting the softmax probability distribution over all 9,714 courses for each student, then removing courses not offered in the target semester and  re-normalizing the probability  distribution. In Fall 2015, 1,937 courses were offered, all of which were candidates for prediction of students' next semester enrollments across all majors, making this prediction problem firmly in the category of large multi-class classification. 
The sub-train/validation set split was also used to choose between 2-gram and 3-gram models. Since the popularity baselines were non-parametric, there was no need to tune them to the validation set. Instead, they were trained on the four Fall semesters previous to the Fall 2016 test set. The RNN model at each hidden node size that performed best on the validation set was re-trained on the entire training set (up through Summer 2016) and was used to predict the test set. In the production version of the recommender system, the model is re-trained in the middle of each semester, after student enrollments have stabilized. The model is again re-trained after the conclusion of the semester when grades are released. 

\begin{table}[htbp]
  \centering
  \caption{Hyperparameters and features explored for prediction of the validation set (Fall 2015).}
    \begin{tabular}{l|l}
  
    \textbf{Hyperparameter } & \textbf{Values} \\
    \hline 
    Hidden layers  & 1,2,3,4 \\
    Hidden nodes per layer & 64, 128, 256 \\
    Entry type, GPA, and major  & With and without each \\
 
    \end{tabular}%
  \label{parameters}%
\end{table}%

We used Recall@10 as the primary evaluation metric for the predictors. This is the percentage of actual enrolled courses that were contained in the model's 10 highest probability course predictions was calculated for each student. We chose to evaluate the top 10 predictions because this is the number of courses which will be shown to  students on the first page of our course information system and is also a common number of items displayed in information retrieval systems (e.g. Google search). The reported recall metric is the average of these percentages across all students predicted. The secondary metric used was Mean Reciprocal Rank (MRR@10). This is the average multiplicative inverse of the \ensuremath{rank_i} of the first correctly predicted course, where $|S|$ refers to the number of students. This is a metric common in information retrieval which focuses on the rank of the first relevant result returned in a search query. The best score of 1 is achieved if the highest probability (first) prediction is always a hit. If the average rank of the first hit were 2, then the MRR would evaluate to 0.50. In the case that a student's first hit is ranked higher than 10, the inverse of the rank for that student is counted as 0. 

\begin{equation*}
Recall = \frac{1}{|S|}\sum_{i=1}^{|S|}\frac{|courses_i \cap predicted_i|}{|courses_i|}
\end{equation*}

\begin{equation*}
MRR = \frac{1}{|S|}\sum_{i=1}^{|S|}\frac{1}{rank_i}
\end{equation*}

\subsection{Experimental Results}
\subsubsection{Results on Validation Set}
In Table \ref{recall10}, we compare four different variants of our basic LSTM model on our validation set. The  results show little variation, with the base course sequence capturing most of the signal and the 256 node LSTM model with major, entry type, and GPA as performing best on Recall, with around 35\% Recall@10, which equates to one third of the top ten recommendations being courses that a student enrolled in that semester. All reported LSTM results used a single hidden layer, as additional hidden layers did not improve prediction with any of the models. 
\begin{table}[htbp]
  \centering
  \caption{Best 5 prediction models based on validation set Recall}
    \begin{tabular}{p{22em}|cc}
    \textbf{Model} & \multicolumn{1}{p{5em}}{\textbf{Recall@10}} & \multicolumn{1}{p{5em}}{\textbf{MRR@10}} \\
    \hline
    256 nodes LSTM + entry type & \textbf{34.62\%} & \textbf{0.5042} \\
    256 nodes LSTM + entry type + GPA & 34.61\% & 0.5040 \\
    256 nodes LSTM + major + entry type + GPA & 34.57\% & 0.5086 \\
    128 nodes LSTM + entry type + GPA & 34.24\% & 0.5007 \\
    256 nodes LSTM + major + entry type & 34.22\% & 0.5064 \\
    \end{tabular}%
  \label{recall10}%
\end{table}%

In Table \ref{averrecall}, we collapse Recall@10 and MRR@10 over each hyperparameter value to show the average performance of that single value across all experiments. From the table, we draw the same conclusion, that higher number of LSTM nodes size give a better result. Adding entry type increases the prediction results but major and GPA do not have a strong influence. It could be that the course enrollment sequence quickly identifies the major a student is in by the subject of courses selected. The utility of GPA may be diminished too because success in a previous semester may be inferable from the selection of courses in the next semester, thought this information would have been expected to benefit next semester prediction in that case. Students are allowed to drop courses late in the semester which contributes to fewer low grades.

\begin{table}[htbp]
  \centering
  \caption{Average hyperparameter value performance on the validation set.}
    \begin{tabular}{c|p{5em}|c|c}
          & \textbf{option} & \multicolumn{1}{p{5em}|}{\textbf{Recall@10}} & \multicolumn{1}{p{5em}}{\textbf{MRR@10}} \\
    \hline
    \multicolumn{1}{c|}{\multirow{3}[6]{*}{\textbf{LSTM nodes number}}} & \multicolumn{1}{c|}{256} & \textbf{34.27\%} & \textbf{0.4986} \\
\cline{2-4}          & \multicolumn{1}{c|}{128} & 33.93\% & 0.4888 \\
\cline{2-4}          & \multicolumn{1}{c|}{64} & 32.72\% & 0.4669 \\
    \hline
    \multicolumn{1}{c|}{\multirow{2}[4]{*}{\textbf{Major}}} & With  & 33.62\% & 0.4806 \\
\cline{2-4}          & Without & 33.66\% & 0.4889 \\
    \hline
    \multicolumn{1}{c|}{\multirow{2}[4]{*}{\textbf{Entry type}}} & With  & 33.89\% & 0.4930 \\
\cline{2-4}          & Without & 33.39\% & 0.4765 \\
    \hline
    \multicolumn{1}{c|}{\multirow{2}[3]{*}{\textbf{GPA}}} & With  & 33.68\% & 0.4864 \\
\cline{2-4}          & Without & 33.60\% & 0.4830 \\
    \end{tabular}%
  \label{averrecall}%
\end{table}%

\subsubsection{Results on Test Set}
After evaluation on the validation semester, Fall 2015, we took the best performing models and re-trained them, keeping their original hyperparameter settings, but using an extended time span of data from Fall 2008 now through Summer 2016. These models then predicted the true test set of Fall 2016. The results are shown in Table \ref{test} and represent the same order of models and same performance magnitude as was achieved on the validation set, suggesting that the model successfully generalizes to unseen semesters.

\begin{table}[htbp]
  \centering
  \caption{Test set prediction performance of the best validation set models.}
    \begin{tabular}{p{22em}|cc}
    \textbf{Model} & \multicolumn{1}{p{5em}}{\textbf{Recall@10}} & \multicolumn{1}{p{5em}}{\textbf{MRR@10}} \\
    \hline   
    256 nodes LSTM + major + entry type + GPA & \textbf{34.59\%} & \textbf{0.5011} \\
    128 nodes LSTM + entry type + GPA & 32.82\% & 0.4787 \\
    64 nodes LSTM + major + entry type + GPA & 32.69\% & 0.4793 \\
     Popularity with major & 23.45\%  & 0.2954 \\
     3-gram & 15.24\% & 0.3090 \\
     Popularity & 8.08\%  & 0.0806 \\
    \end{tabular}%
  \label{test}%
\end{table}%

\begin{figure}
\centering
\includegraphics[height=2in, width=3.8in]{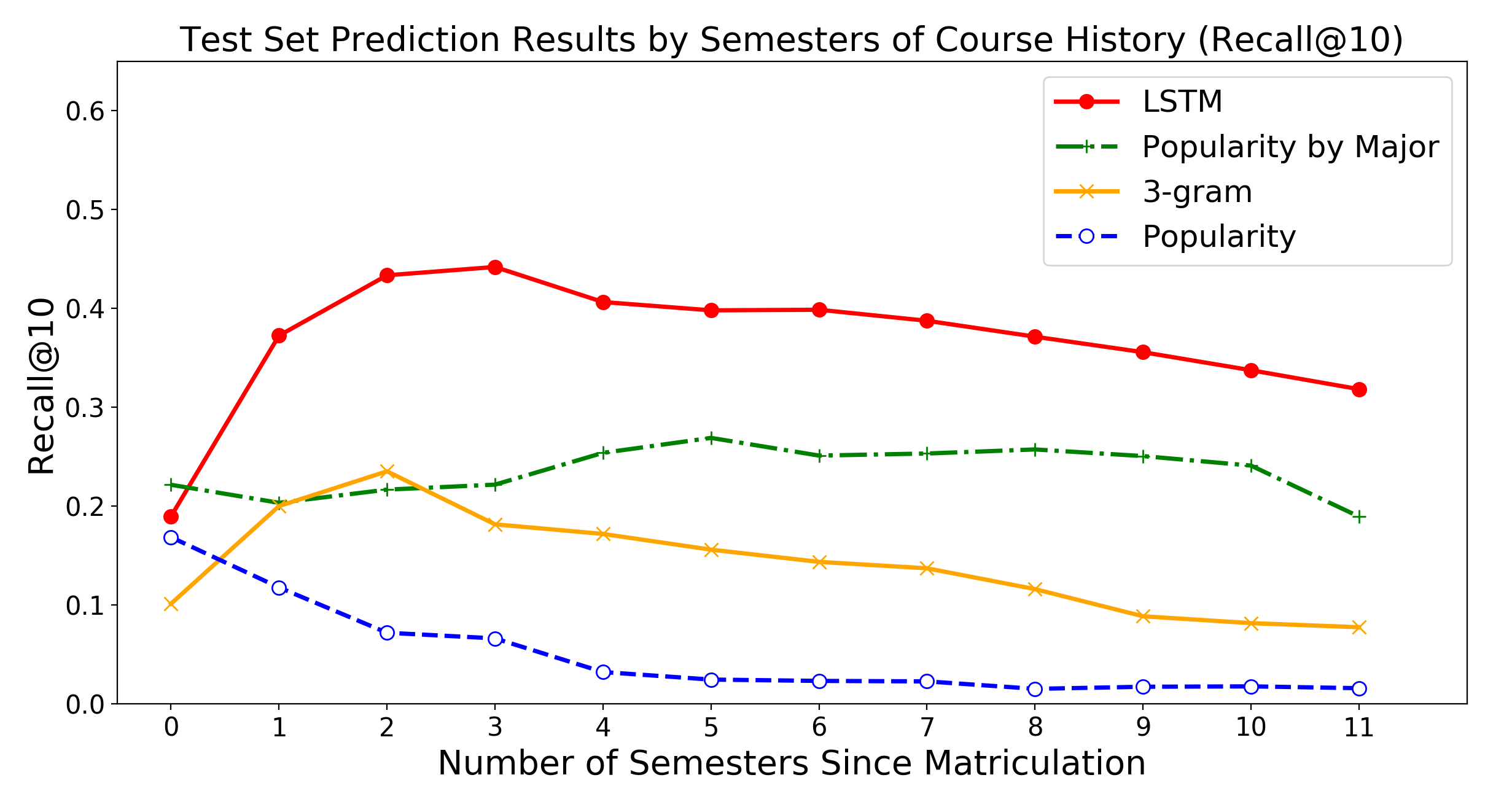}
\caption{Plot showing Recall@10 for the n-gram, popularity, and the best LSTM models on the test set varied by the number of semesters test set students had been at the University prior to the test set semester.}
\label{figure5}
\end{figure}

\begin{figure}
\centering
\includegraphics[height=2.5in, width=4.5in]{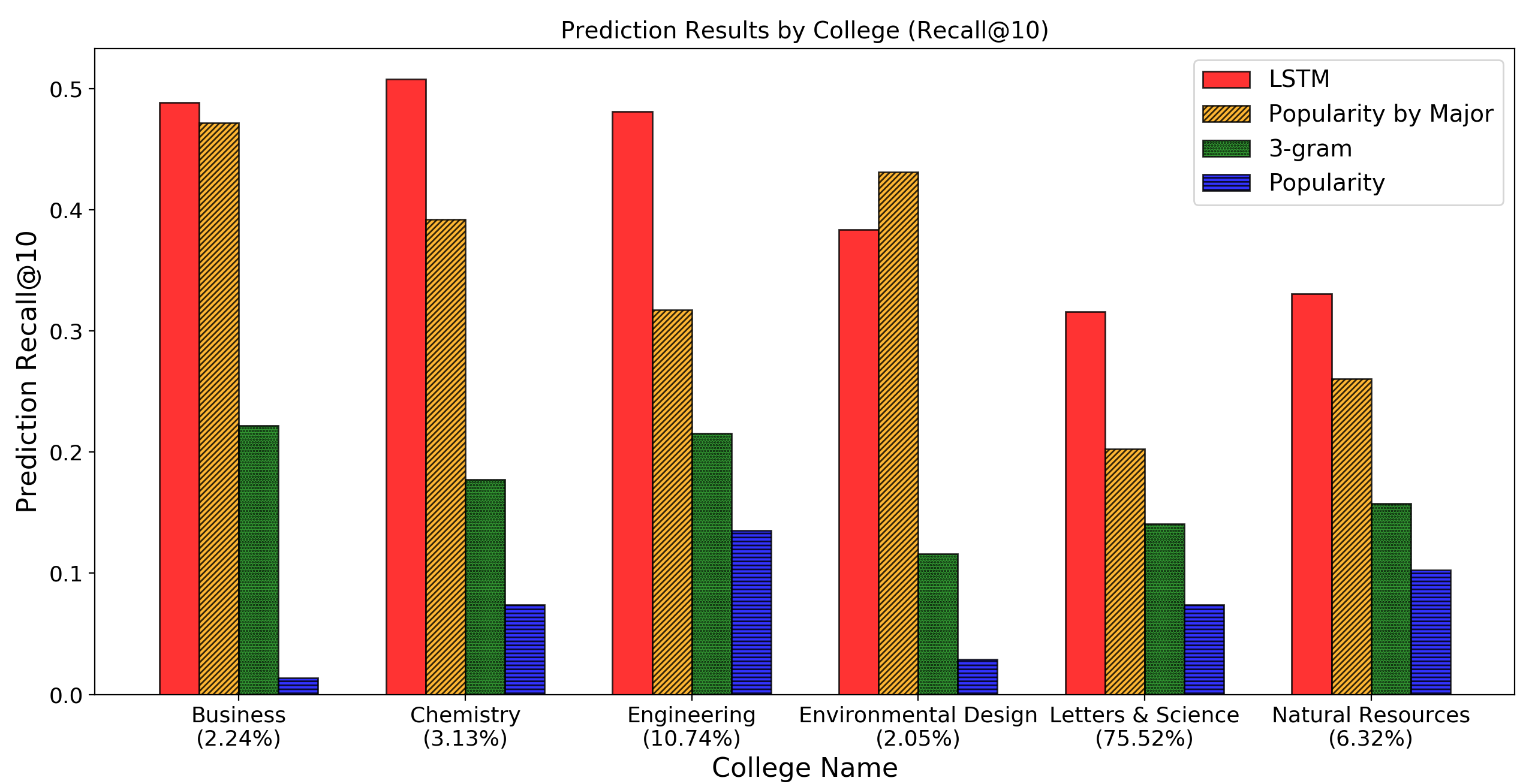}
\caption{Bar plot showing the prediction results (Recall@10) broken out by college with the percentage of students enrolled in a major in that college shown below the college name.}
\label{figure6}
\end{figure}

We plotted Recall@10 based on how many semesters the student had been at UCB previous to the test semester (see Fig. \ref{figure5}). The LSTM differentiates itself most in the second through the fourth semesters of a student's career. In these semesters the LSTM is picking up on a predictive signal that is not as well detected by the other methods. It could be that students are satisfying general requirements at this stage right before degree declaration. We investigated how much the models differed in performance when broken out by the college of the student being predicted. In Fig. \ref{figure6}, we aggregate the recall metric across all majors within the colleges on campus and find that the largest unit, Letters \& Sciences, has the highest percentage improvement of LSTM over the second best model. We also observe that Chemistry had the most predictable course enrollments, with the LSTM correctly predicting half of the students' enrollments on average. The smaller programs in Business and Environmental Design, each of which make up for around 2\% of majors, show that the popularity by major model meets or exceeds the performance of an LSTM. These programs, given their small size, may have a stronger cohort driven design where the incoming class is tight-nit and tend to take the same sequence of courses. 

\section{Interface / Deployment}
We engaged in the development of a standalone web system to surface these features of personalized course recommendation (from the RNN) and course similarity (from the course2vec model) and give students real-time access to this information. This process of development involved the efforts of a campus wide community, including the Office of the Registrar, campus IT, the Office of Equity and Inclusion, the Office of Planning and Analysis, and six undergraduate research assistants. This community effort was a necessary ingredient for a large scale project such as this involving research and practice and is the underlying character of collaboration argued for in the socio-technical vision of learning analytics \cite{siemens2011open}. To make the recommendations real-time, we connected the system to live student enrollment information systems via a campus API which retrieved their course history after logging in using their campus credentials through a common authentication service (CAS). The first incarnation of the system, piloted in the next section, was developed using Sails and Node.js with a Python Flask back-end web service housing the Keras RNN model and gensim \cite{rehurek_lrec} learned skip-gram vectors. Development after the usability study switched to Angular for the front end in order to support unit tests for feature reliability.

\begin{figure}[!tbp]
  \centering
  \subfloat[Course Alternatives]{\includegraphics[width=0.49\textwidth]{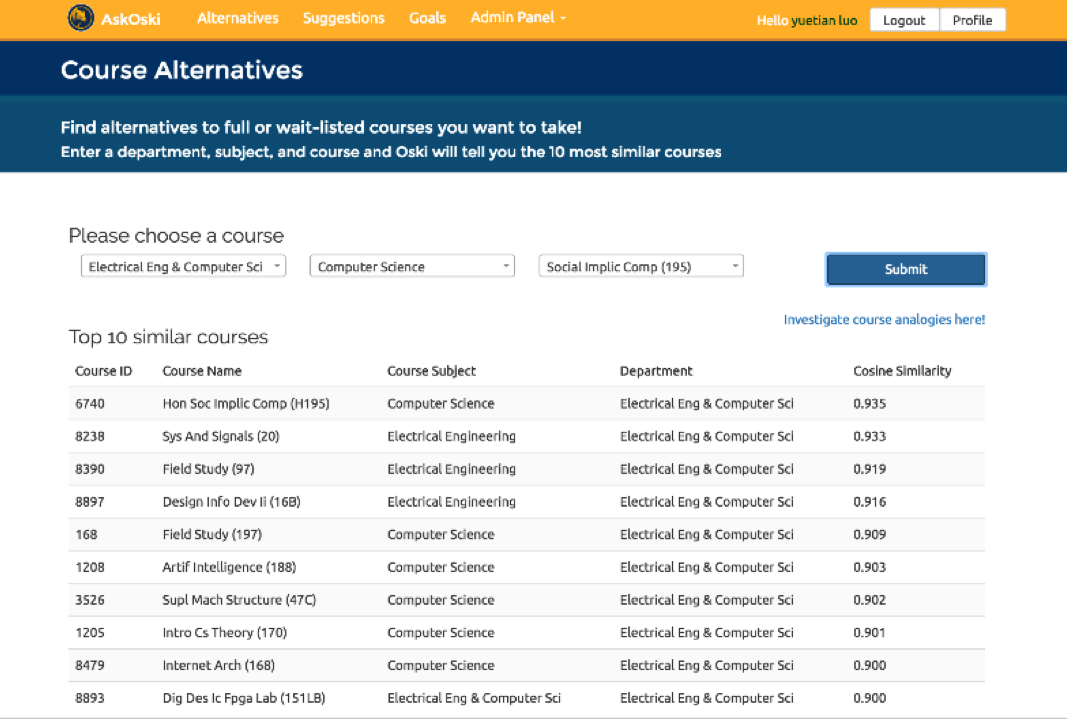}\label{a}}
  \hfill
  \subfloat[Course Recommendation]{\includegraphics[width=0.45\textwidth]{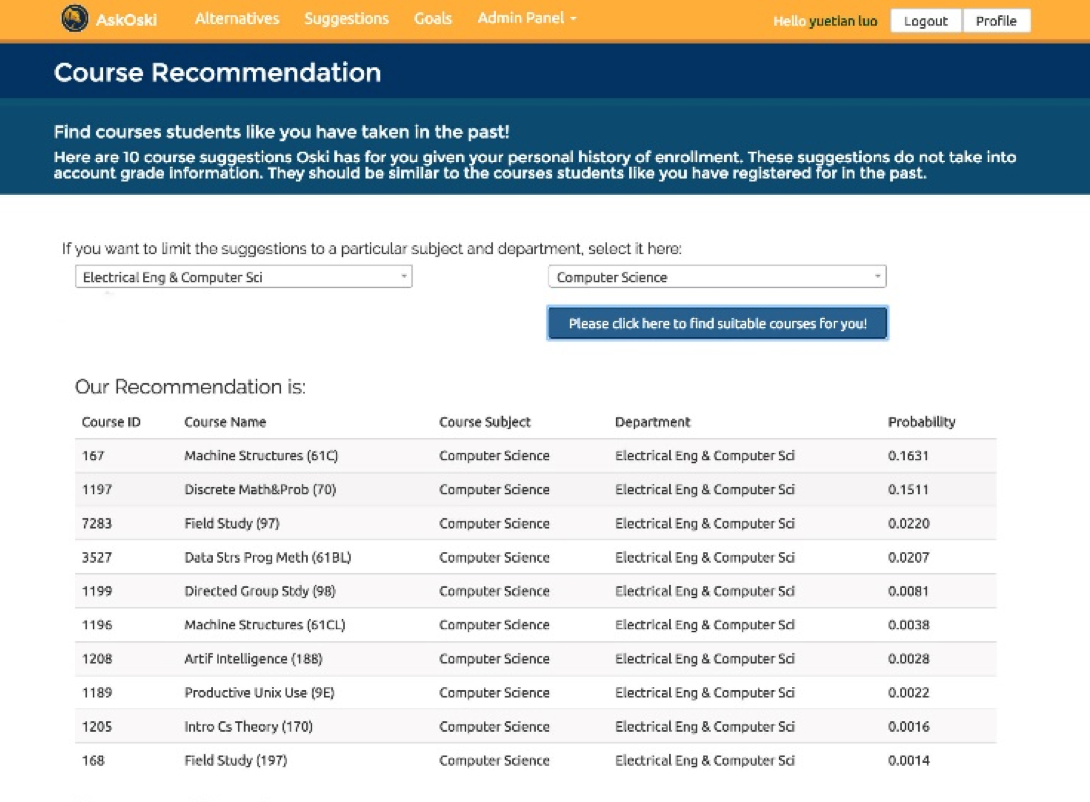}\label{b}}
  \caption{UI depicting the basic functionality of the system during the usability study: (a) course alternatives (b) course recommendation.}
  \label{interface_old}
\end{figure}

\subsection{Usability Study}
\begin{figure}
\centering
\frame{\includegraphics[width=4.5in]{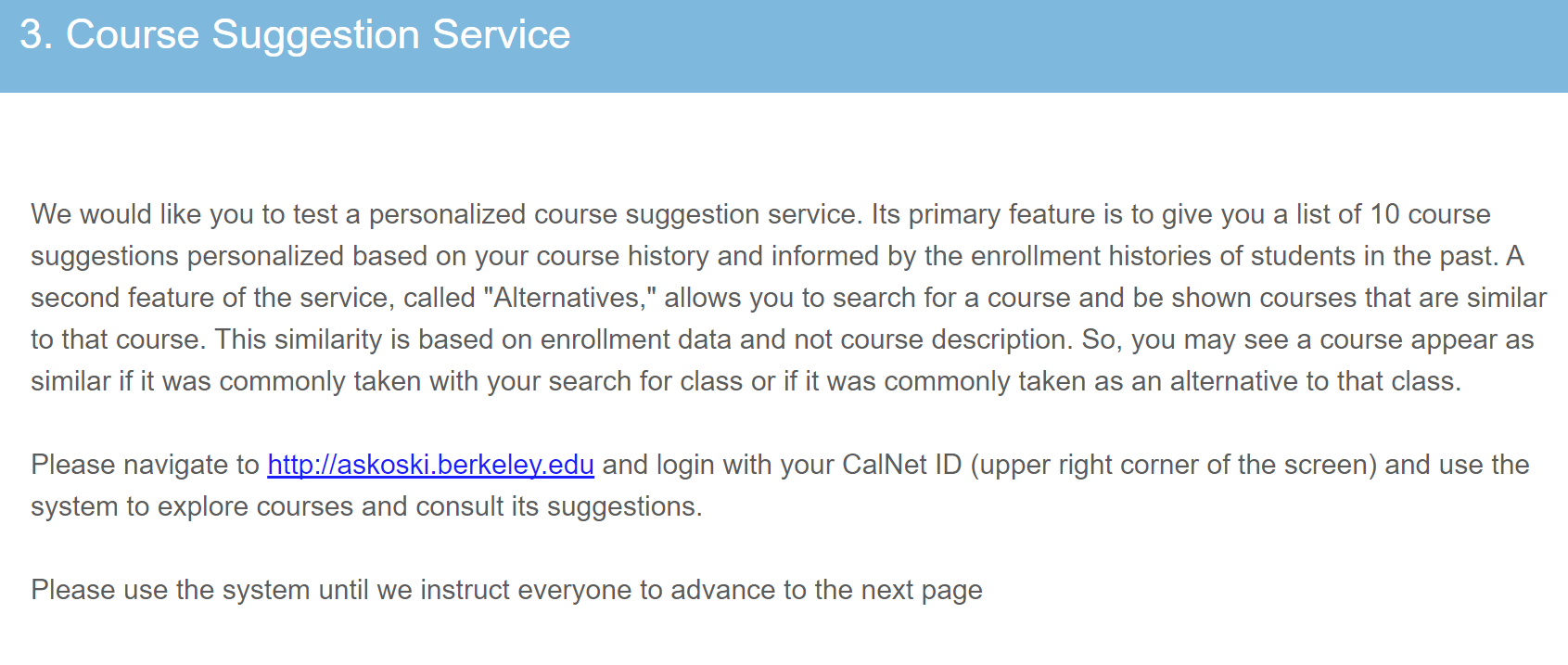}}
\caption{Instructions given to users participating in the usability study.}
\label{survey3}
\end{figure}

\begin{figure}[!tbp]
  \centering
  \subfloat[]{\includegraphics[width=0.57\textwidth]{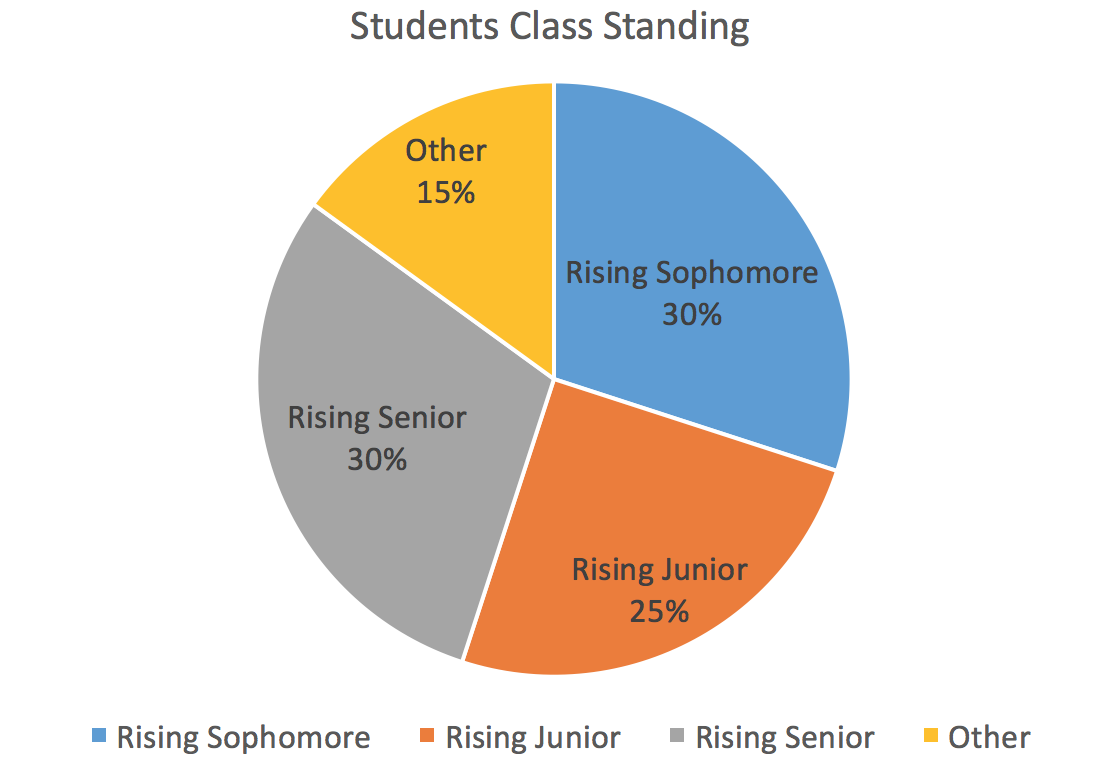}\label{a}}
  \hfill
  \subfloat[]{\includegraphics[width=0.4\textwidth]{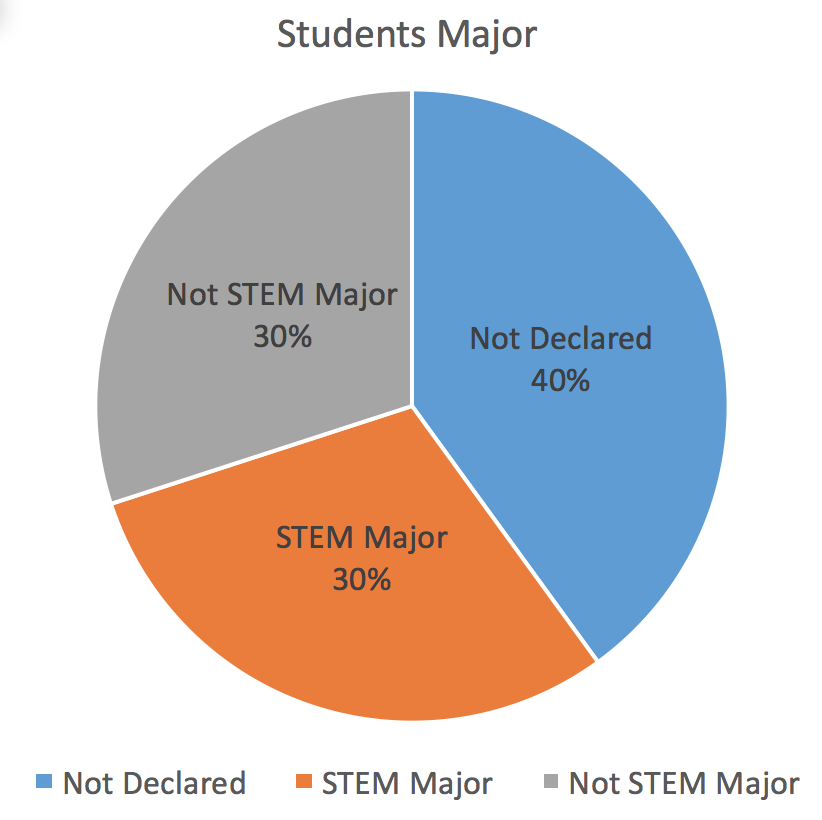}\label{b}}
  \caption{(a) Students' class standing and (b) major distribution. }
  \label{students}
\end{figure}

\begin{figure}
\centering
\frame{\includegraphics[width=3.5in]{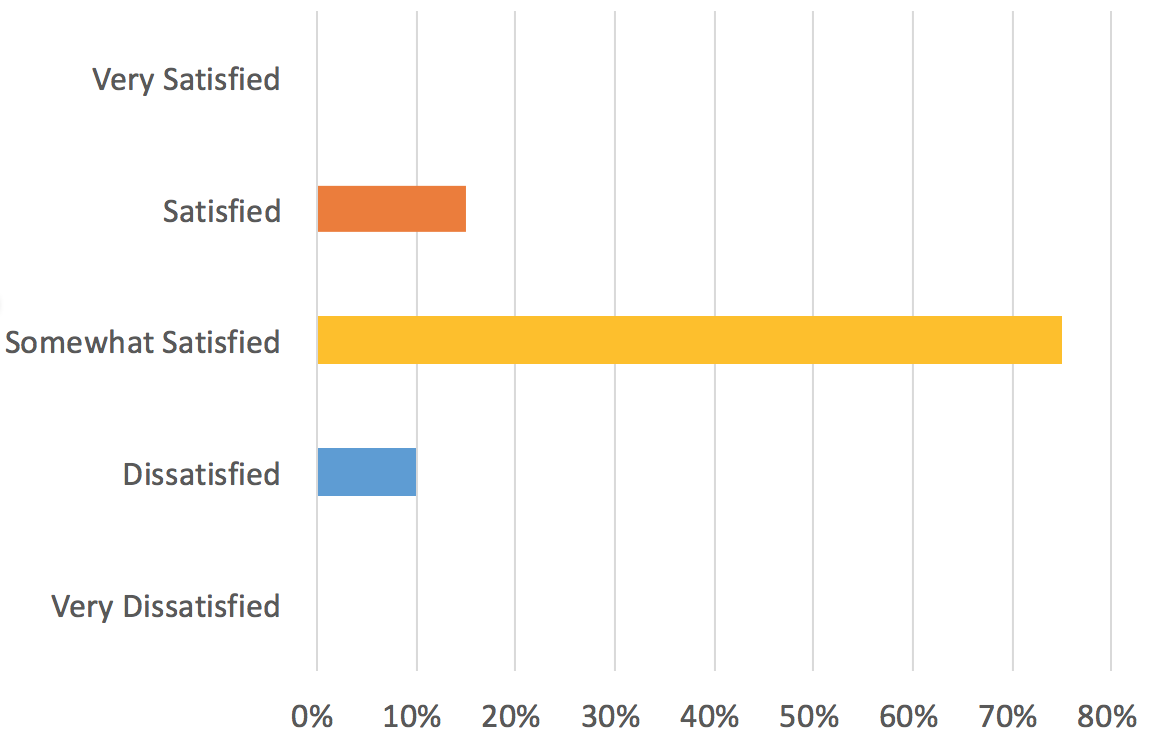}}
\caption{Students' satisfaction with UC Berkeley's current enrollment planning and registration process.}
\label{satisfaction}
\end{figure}

\begin{figure}
\centering
\frame{\includegraphics[height=2.5in, width=4in]{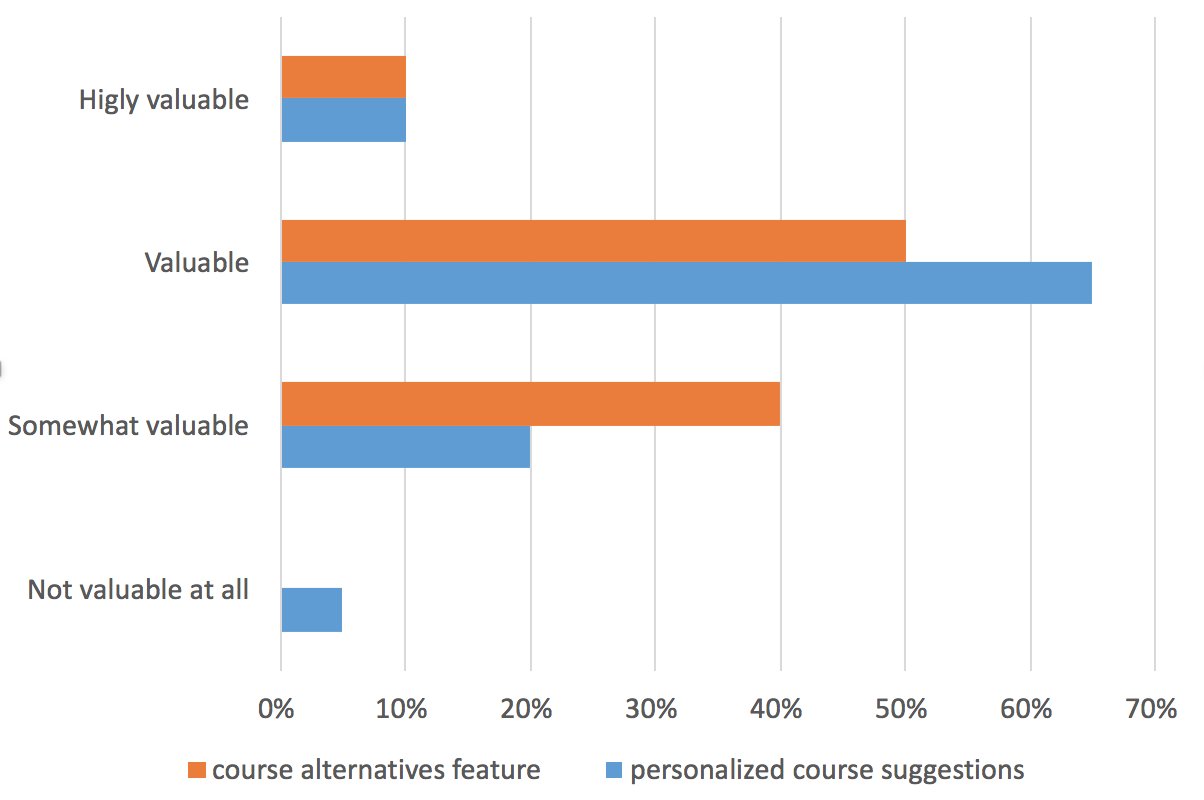}}
\caption{Students' perceived value of the two primary features of the system.}
\label{recvaluebar}
\end{figure}

We believe our collaborative and content-based recommendation design is a novel contribution to research, but, much like most research contributions, its utility to practice was uncertain. We conducted a small scale 20 student usability study in order to measure how well aligned our recommendation approach was to users' needs and to collect feedback on areas of improvement. In an ecosystem where students are gathering information relevant to course selection from numerous sources, we wanted to gauge the background of the intended users and collect feedback on the degree to which they saw our recommendations as novel and useful in the context of that ecosystem. The usability study was conducted in the summer of 2017. A one page flyer was posted on prominent physical bulletin boards around campus and sent to several student mailing lists. The usability test was open to undergrads only and gave three options (Weds, Thurs, Fri) to participate in a one hour session, offering a \$15 gift card to a popular e-commerce website as compensation for their time. The session consisted of six parts:
\begin{enumerate}
\item An introduction of the two proctors of the usability study; the first author and a representative from the UC Berkeley Office of the Registrar. The premise of the session given to the students was that this was a system early in development and that their feedback at this juncture could help shape the direction of the project [5 minutes]
\item Students began section 1 of the survey on computers in the lab in which the session was conducted. This section involved background questions on their class standing and major status (Fig. \ref{students}) [2.5 minutes]
\item Students began section 2 of the survey which asked about their general satisfaction with the existing enrollment experience (Fig. \ref{satisfaction}), their top considerations when selecting courses (Table \ref{considerations}), and the utility of other sources of course information (Table \ref{valuable}) [7.5 minutes]
\item The third section of the survey instructed users to follow a link to trial the recommendation service. The instructions given on this screen are shown in Fig. \ref{survey3} [20 minutes]
\item After twenty minutes, users were asked to return to the survey and advance to the final section, which asked them to rate the usefulness of the two features (Fig. \ref{recvaluebar}), their acceptance of the platform as a whole (Table \ref{acceptanceofrec}), and to give open-ended feedback on ways in which they would like to see the service improved [10 minutes]
\item The last part of the session was dedicated to a retrospective think-aloud \cite{kuusela2000comparison} where users were asked to verbalize their experience with the service. This segment of the session was meant to solicit expanded anecdotes which may have been abbreviated in their text responses and to see if any consensus surfaced with respect to positive and negative experiences with the system [15 minutes]
\end{enumerate}

At the time of this usability study, the only features implemented in the system were displaying the top 10 suggested courses (via RNN) personalized to their course history and showing the top 10 courses most similar to a selected course (via skip-gram). These features as displayed to users can be seen in Figure \ref{interface_old}. In informal testing of the system with undergraduates in the research lab, the suggestion was made to add a filter to the course suggestions which made sure to (1) only show courses offered in the next semester, (2) never show courses that a student had taken before, and (3) never show courses that have a credit restriction (equivalence) with courses the student had taken before. We implemented these basic filters before conducting the usability study.


\begin{table}[htbp]
  \centering
  \caption{Q: How important are the following considerations when selecting classes for the next academic year? Average ratings are on a scale from 1 (not important) to 5 (very important).}
    \begin{tabular}{p{18em}|c}
    \textbf{Student Course Selection Priorities} & \textbf{Average Rating} \\
    \hline
   {Satisfying major requirements} & $\boldsymbol{4.85}$ \\
   
   {Graduating on time} & 4.25 \\
    
   {Maintaining GPA} & 4.15 \\
   {Intellectual enrichment} & 4.10 \\
   {Career goals} & 4.05 \\
   {Satisfying course prerequisites} & 4.05 \\
   {Satisfying breadth requirements} & 3.80 \\ 
    \end{tabular}%
  \label{considerations}%
\end{table}%


\begin{table}[htbp]
  \centering
  \caption{Q: How valuable have the following sources of information and guidance been in helping you make enrollment decisions? Average ratings are on a scale from 1 (not valuable at all) to 4 (highly valuable). A fifth category of ``have not heard of it" comprised 14\% of the ratings and was omitted from the averages.}
    \begin{tabular}{p{25em}|c}
  
    \textbf{Existing Sources of Course Information} & \textbf{Average Rating}\\
    \hline \\[-2ex]
    Berkeleytime\footnote{\url{https://www.berkeleytime.com/}} (grade distributions and enrollment rates) & $\boldsymbol{3.44}$ \\
    {Schedule builder (old version of single semester planner)} & 3.29 \\
    {Ratemyprofessor\footnote{\url{http://www.ratemyprofessors.com/}} (external instructor reviews)} & 3.12 \\
    {Friends / peers} & 3.05 \\
    {Schedule Planner (new version of single semester planner)} & 2.89 \\
    {Ninja courses\footnote{\url{https://ninjacourses.com/explore/1/}} (externally created course catalog)} & 2.83 \\
    {Academic Guide\footnote{\url{http://classes.berkeley.edu}} (new internal course catalog)} & 2.71\\
    {Academic adviser} & 2.24 \\
    {Multi-year Planner (multi-semester planner)} & 2.14 \\
    {Facebook groups} & 1.71 \\
    
    \end{tabular}%
  \label{valuable}%
\end{table}%



The questions in Table \ref{acceptanceofrec} had the Likert scale response options of 
strongly agree: 5, agree: 4, neutral: 3, disagree: 2, strongly disagree: 1. The questions reflected the constructs of a user-centric evaluation framework for recommender systems \cite{pu2011user} designed to assess attitudes and acceptance. In this user-centric evaluation framework, perceived ease of use and perceived usefulness both show significant impacts on overall satisfaction. Moreover, perceived ease of use, perceived usefulness, and overall satisfaction all showed significant impacts on use intention. The average ratings of the statements were all above 3.55 which shows moderately high acceptance of our recommender system among the 20 student usability study group.

\begin{table}[htbp]
  \centering
  \caption{Average ratings of agreement with statements of system acceptance. Ratings are from 1 (strongly disagree) to 5 (strongly agree).}
    \begin{tabular}{p{20em}|p{10em}|p{3.5em}}
    \textbf{Measure of Acceptance} & \textbf{Construct}  & \textbf{Average Rating}\\
    \hline
    - If available, I would use the AskOski service in the future. & Use intentions & $\boldsymbol{4.25}$ \\
    - I would recommend AskOski to a peer. & Use intention  & 3.85 \\
    - The AskOski interface is intuitive to use. & Perceived ease of use & 3.80 \\
    - The AskOski service provided information I couldn't get elsewhere. &Perceived usefulness & 3.55 \\
    - The AskOski service improved my comfort with enrollment planning. &Overall satisfaction & 3.55 \\
    
    \end{tabular}%
  \label{acceptanceofrec}%
\end{table}%

\begin{table}[htbp]
  \centering
  \caption{Representative quotes from open-ended feedback on ways students would like to see the service improved.}
    \begin{tabular}{p{35em}}
    \textbf{Quotes (student feature requests)}\\
    \hline
    \begin{enumerate}[]
    \item \textit{Explanation as to why the classes are similar to the ones the student has taken }
    \item \textit{Would like AskOski to be able to create a visual instead of just a table of courses and similarities, maybe like a mind-map with colors would be nice.}
    \item  \textit{Wish the suggested courses were linked to descriptions of the suggested courses so that we could verify for ourselves how similar these courses are.}
    \item \textit{User profile: Allow more than 1 major in profile.} 

    \item \textit {Having to select a department and a subject is somewhat redundant.}
    \item \textit{To see how close we are to completing major prereqs. Since the system has our major and our class history, that checklist would be nice.}
	\item \textit{Could just select a subject/department and just see a list of classes in that department. Sometimes people don't know all of the classes they want to take.}
    
    \end{enumerate}
   
     \end{tabular}%
  \label{open-ended}%
\end{table}%









\subsection{Integration of User Requested Features}
\begin{figure}
\centering
\includegraphics[width=\columnwidth]{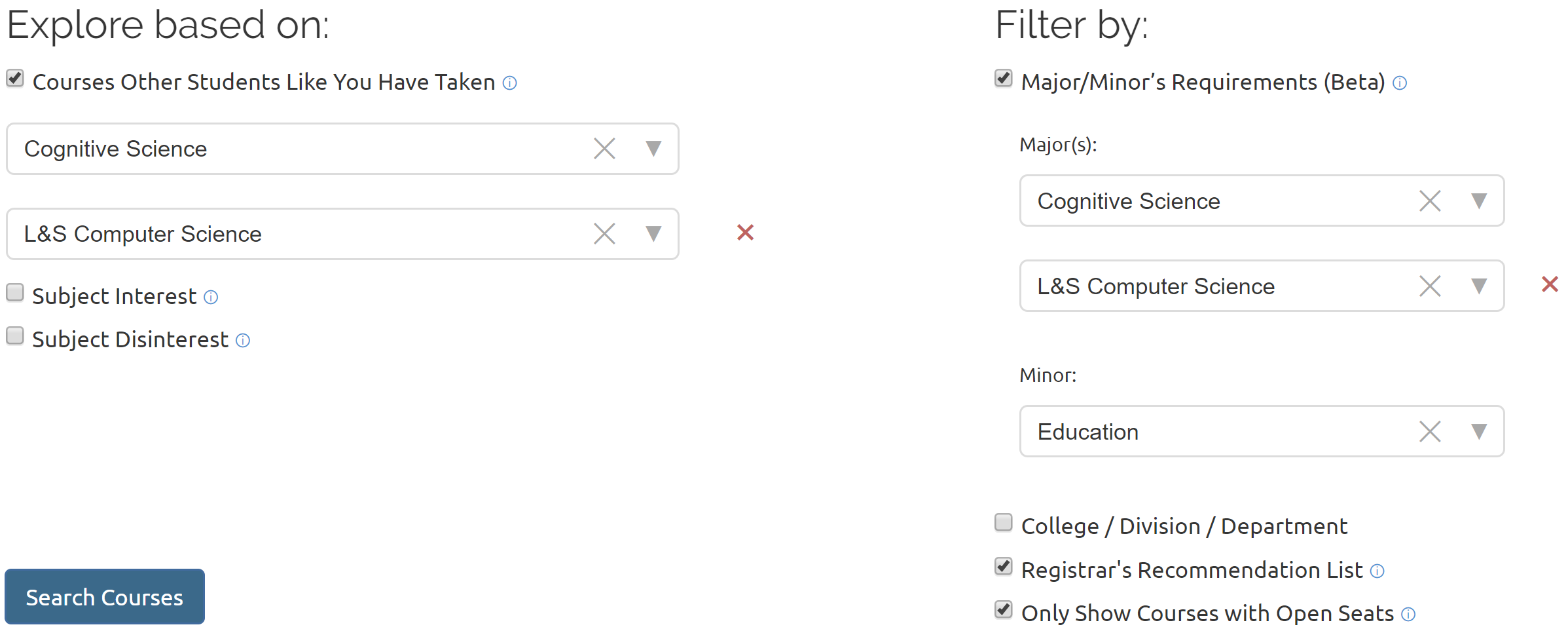}
\caption{Screenshot of the options portion of the recommendation system implemented in response to user feedback. Options which affect the sorting of courses are on the left, while options which restrict what is being shown are on the right.}
\label{options}
\end{figure}

Takeaways from the usability study were that (1) students highly prioritize satisfying degree requirements (2) students generally accepted the features of the service as useful and (3) students are seeking information about courses from a wide variety of sources, least of which is their academic adviser. We took into account the suggestions of students made in the open ended responses (Table \ref{open-ended}) and implemented most (3,4,5,7), with suggestions 1 and 2 being a subject of research covered in the next section of this paper. Figure \ref{options} shows the new options interface which reflects the features added in response to user feedback. The left section of the options display is labeled ``Suggestions based on:" and allows the student to select one or more majors they have declared in order to receive personalized course considerations for the next semester. These majors are automatically filled in via a student API request that reports back the major(s) of the student but can be changed by them at any time. Students who have not yet declared a major can enter in different majors they are considering in order to explore what potential future pathways look like. Students can also completely turn off the collaborative recommendation bias. In the case that no boxes are checked, courses will be displayed alphabetically by department and then by course number. The interface, without collaborative bias turned on, can be quickly personalized by the student by utilizing the ``Filter by:" options on the right. These allow students to filter the displayed courses by (a) courses that satisfy one or more of their major(s) or minor (b) by a college, division, or department (c) courses on the registrar's recommended list or (d) courses with open seats (updated daily). Any number of these filters can be simultaneously checked in which case they act conjunctively. In its current state, the degree requirements filter is not a personalized list of courses that will satisfy a yet unmet degree requirement of the student. Instead, in its first phase of implementation, it is simply a list of courses that satisfy some requirement of the student's major(s), that will be included regardless of them having satisfied that requirement. The combination of this filter with the sorting by personalized collaborative bias, might result in course recommendations for requirements the student still has not satisfied. An upgrade to this filter will utilize the Academic Plan Review (APR) campus system that will enable  fully personalized degree requirement tracking, request number 6 in Table \ref{open-ended}.  

Recommendation is a multi-stakeholder activity \cite{abdollahpouri2017recommender}. In the context of the university, it involves not just the students as stakeholders but those affected by their course selections, such as the faculty who teach the courses, the administration that prepares the scheduling based on anticipated enrollments, and the allocation of teaching assistants. Under-enrolled classes are advertised by the registrar on an embedded Google spreadsheet in a section of their website. We import this list, updated daily, so that students may have it presented to them in a friendlier interface and also have the option of personalizing this 100+ class list according to the interests suggest by their personal history or by explicit preference specification with the subject interest and disinterest feature, described in the next section. By joining in an external list with our existing model infrastructure, we can bring new life to this or any other collection of courses from the catalogue. 

\subsection{Integration of Explicit User Preference into Recommendations}
\begin{figure}
\centering
\includegraphics[width=\columnwidth]{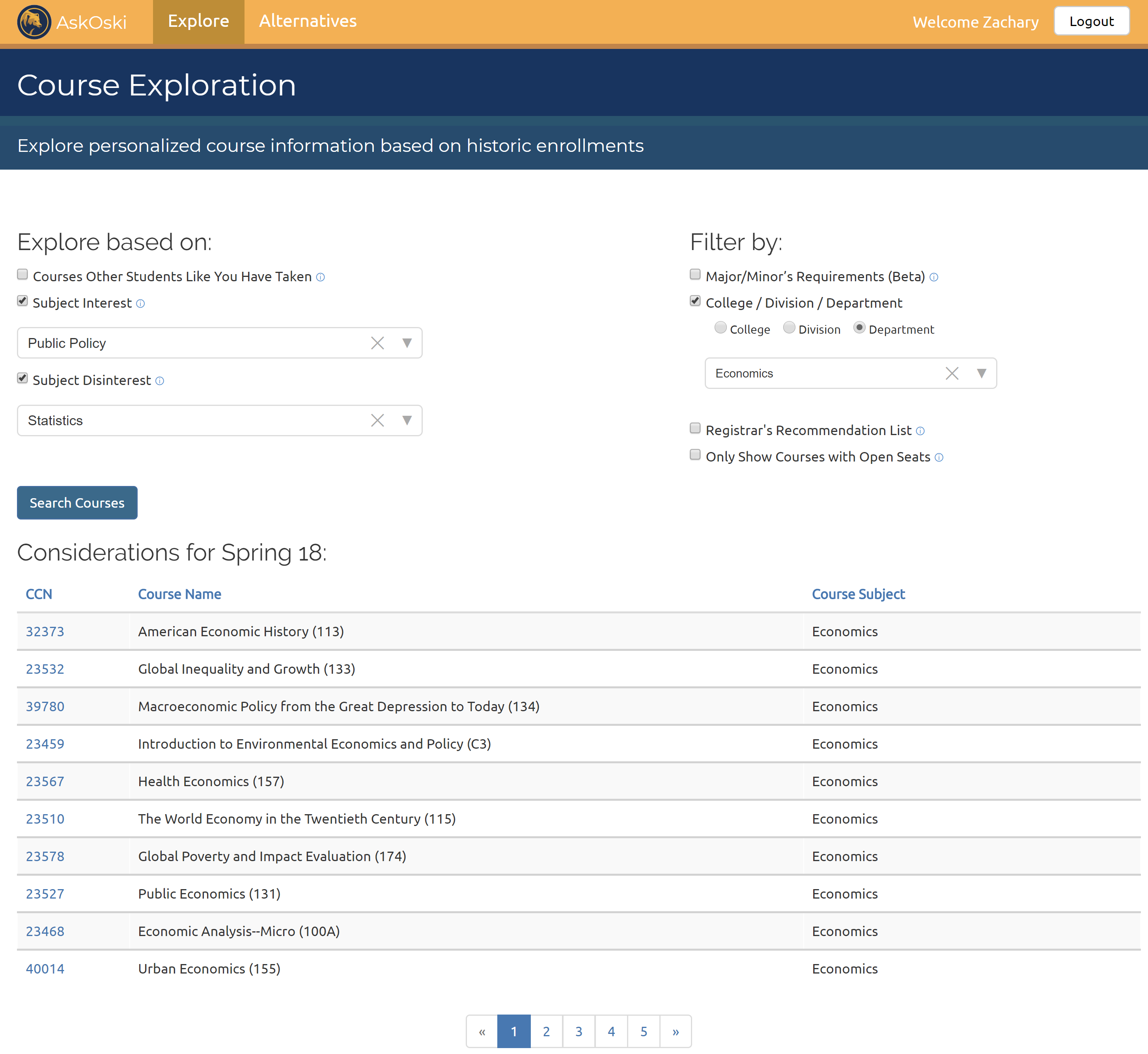}
\caption{Screenshot of the suggestion service highlighting the subject interest and disinterest feature. In this example, Public Policy is selected as the subject interest and Statistics as the disinterest with a filter on only courses within Economics. The result of this query is a sorting of courses maximizing their closeness to Public Policy and distance from Statistics in the course vector space.}
\label{newversion}
\end{figure}

We take into account student explicit preferences by leveraging the implicit semantics of the course vector space. In this feature, we respond to the combination of a student request from the usability study survey (7), that courses be allowed to be filtered by department, with a sentiment that emerged in the think-aloud session, that students would like to express likes and dislikes. The subject interest and disinterest feature, shown in Figure \ref{newversion}, allows students to sort courses by their inferred relevance (or irrelevance) to specified subjects. In the example depicted, a user chooses to only have courses in the Economics department be displayed. He also specifies a preference for the subject of Public Policy and a disinterest for Statistics. Economics is a highly statistical social science, and therefore its courses often run the gamut of relevance to these two subjects. The result is a list of courses in Economics sorted by courses which are close to Public Policy and far from Statistics. This list is produced by calculating the score $s$ of each course $i$ in the Economics, defined by:

$$s_i = - ||\boldsymbol{v}_i - \boldsymbol{v}_a||'_2 + ||\boldsymbol{v}_i - \boldsymbol{v}_b||'_2$$ 
In the above equation, a statistical bias is created towards the subject interest $\boldsymbol{v}_a$ (Public Policy) and against the disinterest subject $\boldsymbol{v}_b$ (Statistics). Vector representations of these subjects are generated by taking the average (or centroid) of all of their respective course vectors. The term $\boldsymbol{v}_i$ is course $i$'s vector learned from the course2vec model and $||\boldsymbol{v}_i - \boldsymbol{v}_a||'_2$
is the normalized Euclidean distance between course $i$ and the subject interest vector. If the collaborative recommendation option is simultaneously selected, the probability distribution of courses predicted to be taken next by the RNN will be a third term in this equation.

Returning to the example in Figure \ref{newversion}, the top results of the sorted courses by score show face validity, with two courses containing the word ``Policy" in their titles, and the first two courses, on American Economic History and Global Inequality and Growth, both relevant to policy concerns. If these interests are swapped; the top four courses are Introduction to Mathematical Economics, Econometric Analysis, and courses in Macro and Micro Economic Theory. This exemplifies the way in which the vector space model synthesizes features of courses learned from behavioral data and allows for the relationships between any arbitrary object and another object or identified concept to be calculated and surfaced. The results of this type of automatic relevance rating can be seen as a type of information retrieval. Similar to a search engine, 100\% of the top results need not be relevant in order for the service to have utility to the user and this vector approach will certainly not always perform ideally. This querying of how a course relates to a concept or subject can be seen as a desirable scrutable property of a vector space model.

\section{Scrutability}

Vector space models already provide a degree of scrutability in their ability to explain the semantics of an embedded element in terms of its relation to other objects or concept vectors identified in the space. RNNs have less inherent interpretability due to their non-linearities. Student from our usability study reported a desire to know why the system was making its recommendations. In this section, we propose an augmentation to our course RNN model allowing for a moderate level of scrutability of the recommendation by the user. Our proposed model integrates course descriptions to provide users with the top keywords associated with their RNN hidden state, giving them a glimpse into what the system believes is relevant to them and helping rationalize why it is producing the recommendations they see. 
We add an additional analysis, giving a view into similarity of student course pathways by major by applying t-SNE dimensionality reduction \cite{van2014accelerating} to the hidden states of the RNN, a technique increasingly applied to neural networks in a variety of domains \cite{mnih2015human,wang2017learning,pardos2017analysis}.
\subsection{Background on Scrutability}
Scrutability was introduced as a concept in user modeling and intelligent tutoring systems by \citet{kay2000stereotypes}, where it was defined as enabling learners to directly examine what the system believes about them and, to a degree, examine how the system arrives at its beliefs about other users inferred to be different from them. Subsequent work extended the definition of scrutability, specifically as an attribute of a model, as being able to be both examined and modified by the user \cite{czarkowski2000bringing, kay2006scrutability, czarkowski2005scrutability, kyriacou2009evaluating, wang2018tem}. In rule-based models or models where inferences of user attributes can be easily discretized (e.g. estimation of skill mastery), allowing the user to examine the system's beliefs about them is a policy decision posing minimal technical challenge. In the case of neural networks used for user modeling; however, discrete inferred attributes are often not explicit, as is the case with most RNN applications, and thus it is less straightforward to identify what beliefs the system holds that could be in turn scrutinized by the user. In this section, we address this technical challenge of scrutinizing neural networks by mapping the abstract hidden state of the learner to semantics sourced from the descriptions of their selected courses. This mapping of a hidden state to semantics can be seen as analogous to work mapping images to text for the task of auto-captioning \cite{xu2015show}. Explainability has emerged as an adjacent concept to scrutability in recommender systems \cite{KoukiSPOG17}, often presenting a descriptive cause and effect rationale for why a recommendation is being made (e.g. this course is being recommended because you took Linear Algebra last semester). Our contribution to scrutable neural network models in this section pertains to the definition of scrutability introduced by \citet{kay2000stereotypes} of allowing for the examination of what the system believes about the user (i.e. preferences) and how they are modeled in relation to other users of the system.

\subsection{Adding Semantics to the Hidden States of an RNN}
Tagging is a means for adding semantics to an item. With the subject interest/disinterest feature of the recommender, we added semantics to the vector space by using the subject area of a course as its tag and synthesized a vector for that subject which could then be used to generalize its relation to courses outside of the subject. In this section, we demonstrate how semantics can be associated with the hidden state of an RNN. While these states are not vector spaces, useful topical regularities may be produced that do not correspond directly to words in the courses being recommended but nevertheless provide a relevant conceptual description of the current state of the user model. 
We augmented the best model as predicted on the test set (from Table \ref{test}) by adding an auxiliary output of the combined bag-of-words of all the courses taken in that semester by the student. This combined BOW was a multi-hot of the same dimensionality as described in the course equivalency validation of section 4.2 where a 1 was placed in a dimension of the BOW if the corresponding word appeared in any of the courses taken by the student that semester. The auxiliary output layer used a softmax activation with categorical cross-entropy loss, much like the output layer of a skip-gram model. This layer existed alongside the standard course multi-hot output of the existing model. The weighting of each output layer can be adjusted, and so we tuned the weight of the BOW output such that the magnitude of its loss was on the same order as the loss produced by the course enrollment multi-hot. While a sigmoid with binary cross-entropy could be used in place of the current activation and loss, it would produce overly optimistic accuracy due to being rewarded for predicting the highly abundant 0 class representing that a word does not appear in the BOW. We trained the model, this time with the course description BOW auxiliary output and then selected one student at random from the majors of Cognitive Science, Economics, and Chemistry, and recorded the probability distribution over the BOW at each semester. The top 5 most probable words in each semester were logged, with the top words for each student's Fall semester (from their Freshman to Senior years) shown in Table \ref{bow}. Checking the performance of this model on the test set course prediction task, we found that course prediction accuracy was not diminished by the addition of this output (Recall@10 improved by an insignificant margin).

\begin{table}[htbp]
  \centering
  \caption{The five most probable words from the BOW output of the LSTM describing the hidden state representation of three students in the majors of Cognitive Science, Economics, and Chemistry in the Fall semester of each of their four academic years. The keywords represent an interpretation of the model's belief in what is of relevance to the student at each semester.}
    \begin{tabular}{c|l|l|l|l}
          & \textbf{Freshman} & \textbf{Sophomore} & \textbf{Junior} & \textbf{Senior} \\
    \hline
    \multicolumn{1}{c|}{\multirow{5}[2]{*}{\textbf{Cognitive Science}}} & major & language & relation & social \\
          & read  & introduction & symbol & examine \\
          & write & science & problem & relation \\
          & requirement & structure & cognition & well \\
          & discuss & logic & psychology & politics \\
    \hline
    \multicolumn{1}{c|}{\multirow{5}[2]{*}{\textbf{Economics}}} & first & determinant & analysis & problem \\
          & equation & allocation & economics & analysis \\
          & differentiate & space & application & theory \\
          & order & application & estimation & model \\
          & application & resource & introduction & introduction \\
    \hline
    \multirow{5}[1]{*}{\textbf{Chemistry}} & radioactive & chemical & molecular & design \\
          & equivalent & application & mechanics & computation \\
          & emphasis & mechanics & kinetics & application \\
          & science & physics & structure & principle \\
          & kinetics & introduction & emphasis & logic \\
    \end{tabular}%
  \label{bow}%
\end{table}%

\subsection{Visualization of student course pathways}
\begin{figure}
\centering
\includegraphics[width=\columnwidth]{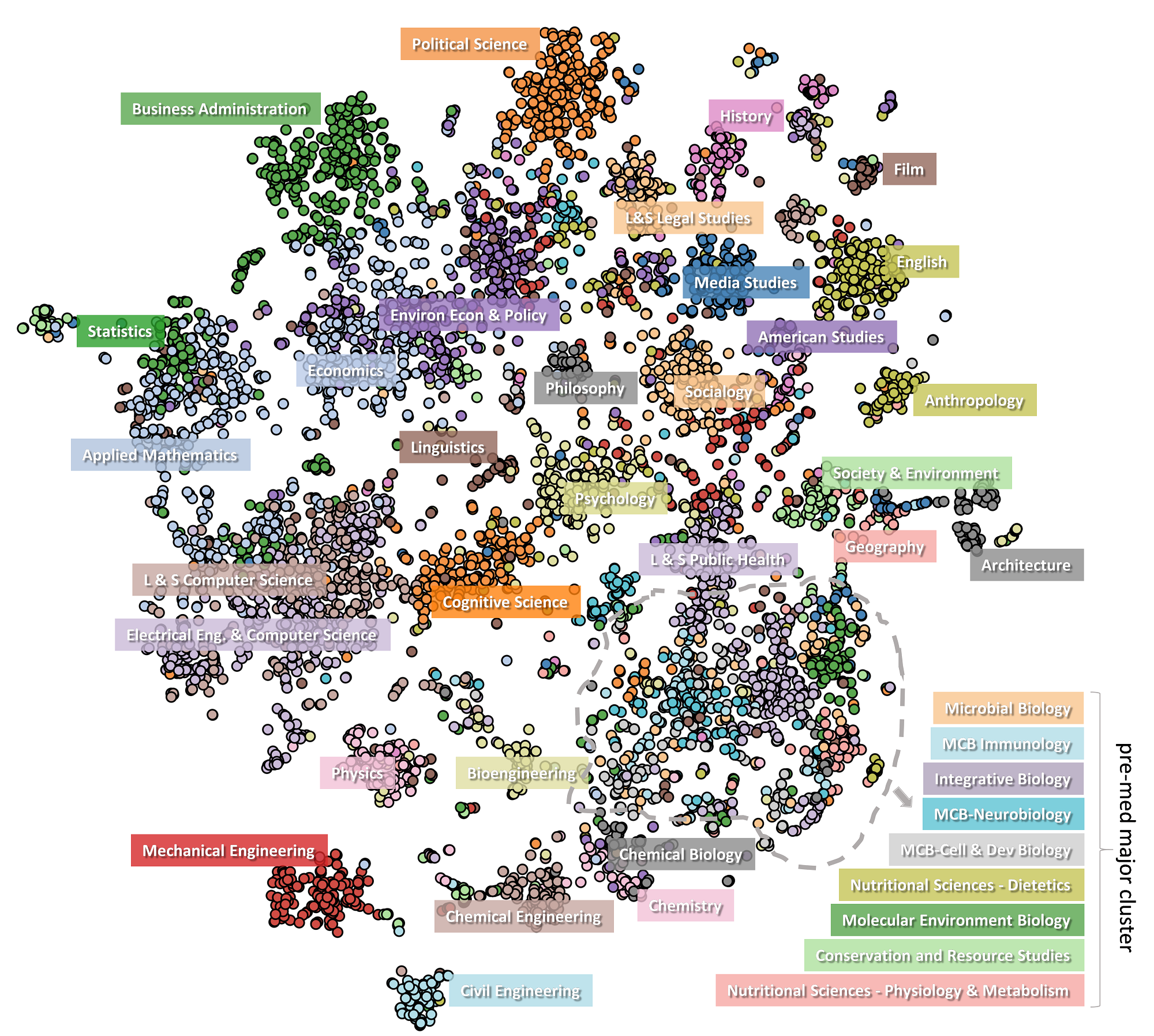}
\caption{Visualization of undergraduates based on their hidden state extracted from the RNN at the beginning of their final academic year. The position of a student in this scatter plot can be seen as summarizing their history of course enrollments.}
\label{tsne}
\end{figure}

We visually inspected our model to see if it would capture the course enrollment similarity and diversity of Berkeley students as they made their way through to the last year of their undergraduate career. Specifically, we considered all 6,103 students that graduated on time with Fall 2016 as their last academic year of study. That included 4,093 new freshmen who matriculated in Fall 2013 and 2,010 transfers who matriculated in Fall 2015. We fed their enrollment and major data through our best performing model from Table \ref{test} and extracted the hidden state of the LSTM for each student at their Fall 2016 semester. This hidden state consisted of the first 256 elements of the input of the last fully-connect layer, which did not include the concatenation with the additional features such as major and GPA. We then used Barnes-Hut t-SNE \cite{van2014accelerating} to reduce the dimensionality of the 256 length hidden state to a two dimensional space. t-SNE dimensionality reduction prioritizes retaining closeness in the high-dimensional space in the lower-dimensional space and is a non-linear manifold projection. An interactive d3 scatter plot tool\footnote{\url{https://github.com/CAHLR/d3-scatterplot}}, developed by the research lab for representation cluster analysis, was used to visualize the 2D mapping, coloring the data point of each student based on their major. The results are depicted in Fig. \ref{tsne}. This visualization represents a summary of students' course enrollment paths from matriculation up until the first semester of their last year and suggests which majors followed a similar theme of course selection. 

The majors of Cognitive Science and Psychology can be seen at the very center of the plot, serving as the conceptual linchpins for the hemispheres of STEM and non-STEM majors, respectively. The other boundary majors that can be seen sitting between the mathematical disciplines (e.g. Engineering and the physical sciences) and the social sciences and humanities are; Business Administration, Economics, Linguistics, and Public Health, with the collection of majors often declared in preparation for a medical degree located between Engineering and Public Heath.  

\section{Discussion of Bias}
A model of adaptive personalization can at times come to wrong conclusions, whether constructed from inferences from ratings or behaviors. User choice has been a proposed mediator of potential negative bias in such systems \cite{kay2000stereotypes} and is one strategy employed in our implementation to combat cases where a student's interests do not align with what has been inferred by the model from their course history. Subject preferences can be specified, majors can be added and removed, and the collaborative bias can be turned off completely. Four other strategies, not yet implemented, can also be brought to bear to address gaps in effective collaborative-based recommendation systems. The first is to impose policies or rules with the intent of discouraging historic behavioral patterns an institution wishes to counteract or curtail. This constitutes the introduction of bias in itself, albeit one in the direction an institution or representative governing body deems to be in their interests or the interests of their constituent community. The second alternative strategy is to counteract bias at the model representation level. Word embeddings have scored high on measures of \textit{implicit bias} \cite{bengio2003neural} and have been found to contain negative gender-based stereotypes in analogical associations of gendered words and professions. Approaches have been proposed for modifying embeddings to remove such undesirable associations \cite{bolukbasi2016man}. Since embeddings are trained on members of society's own writings or behaviors, they are a candid reflection of our historic dispositions and should remain unmodified if being studied for anthropological purposes; however, as societal norms are contested and drift, associations may quickly become antiquated and we may like to quickly bring them up to speed if they are being used formatively. The third strategy is therefore to bias towards more recent data in the training of these models so as to reflect modern trends over more historic ones. Lastly, any source of guidance, human or algorithmic, can be prone to exhibiting unintended negative biases and, like with humans, mechanisms should exist for users to provide feedback when they believe a misunderstanding has occurred and perhaps allow the error in representation to be explicitly corrected \cite{bull2010open}. 

Still, in spite of strategies to prevent it, users may be confronted with recommendations from an inaccurate model of them. It is therefore an important question for designers of recommendation systems, particularly in educational settings, to ask under what circumstances does a user's membership in a stereotype turn from a help, to a hindrance, to a harm and what factors influence someone's propensity to be negatively affected by perceived inclusion in an inaccurate or over general stereotype or category. The socio-cultural factors around algorithmic recommendation in formative contexts are no less complex than those faced by human recommenders, such as academic advisers, who at times need to give advice that an advisee may not want to hear.

\section{Limitations}
Several representational limitations exist in the RNN and skip-gram models. For courses that have been offered for many years, their concepts may have drifted; however, we decided to only use a single representation of the course, which likely represented the concepts present in the majority of offerings of the course and not necessarily in the most recent offering. Some courses are taught by one of several instructors. The instructor teaching the course in a particular semester or section can have a significant impact on its material and pedagogy, though this difference will not be reflected in the course description and is also not reflected in our representations as instructor effects were not incorporated into the models. While instructor interest/disinterest would be a popular feature among students, it would likely be a politically untenable one. A more technical limitation of associating an instructor with a course is that our multi-hot input representation paradigm inhibits this association without exponentially increasing the input vocabulary size. In order to associate any feature with a course (i.e. grade, instructor, or year), separate tokens of each course concatenated with each possible feature value must be created. If one course per time-slice were presented, thus compromising the semester structure, the exponential expansion could be avoided by concatenating feature one-hots together without any ambiguity as to which course they were being associated with. In order to add features to the skip-gram representation, a modification of the c++ word2vec implementation \cite{rehurek_lrec} would need to be made, or modeled in a separate extensible framework. This is feasible but in both cases would likely not be without a hit to performance, given the substantial optimization to hardware of the canonical models.

The recommender system is limited by the cold-start problem in two regards. First, when a student is in their first semester at the University, the system currently has no data about the student with which to adapt to. This means that, prior to the user specifying filters or an intended major, the same recommendations would be shown to all first semester students. This problem could be addressed by integrating information about the advanced placement or community college credits the student has earned. Alternatively, a first year student could be scaffolded to specify several candidate majors before suggestions are shown so that the student can be shown first year courses that may be advantageous for pursuing either major. The second cold-start limitation is in the courses available to be displayed. If any of the ``Explore based on" options are selected (the collaborative option is selected by default), the system sorts courses by their collaborative-based probability or in combination with the course's relationship to specified subjects of interest or disinterest. This means that only courses with a vector representation can be sorted, which precludes courses being taught for the first time. To overcome the cold-start issue with new courses, a filter could be introduced which shows only new courses in a random order, or a mapping function can be learned, bootstrapping the course into the RNN embedding and skip-gram vector space by way of its catalog description.

\section{Contribution} 
In this paper, we brought to bear the state-of-the-art in connectionist approaches to collaborative filtering and representation, tuning and validating individual models and integrating them into a personalized adaptive course information and recommendation system deployed at scale, serving students pursuing 265 majors across the university. 

Below, we summarize the work's contributions to areas of representation learning, recurrent neural network design, recommender systems, and the application area of higher-education.

Using a skip-gram for representation learning, we demonstrated how a collaborative based representation of item content could be constructed implicitly from behavioral data. On the task of predicting which courses had high subject matter overlap, representations learned from enrollment sequences performed better in terms of average predicted similarity than a bag-of-words representation of the courses based on their semantic catalog description. When using median instead of average similarity rank, the BOW approach provided better predictions, with a median rank of 4 compared to the skip-gram model's 18 (out of 3,939 candidate courses).

In order to adapt the canonical recurrent neural network design to our application area, we introduced the use of multi-hot input to represent multiple elements (courses) of the same type being observed in the same time slice (semester). The RNN was able to find significant signal in historic enrollments, predicting on average one third of courses selected by students in the test semester, an improvement of 47\% over a popularity by major baseline. The best RNN model used the highest node count in our hyperparameter search of 254 and incorporated all the available features of major, entry type (transfer/new freshman), and GPA of the previous semester. We added a degree of scrutability to the RNN by creating an auxiliary BOW output of the descriptions of courses taken by a student in the predicted semester. By extracting the highest predicted words at each time slice, a glimpse into the semantics behind the model's inferred areas of relevance to the user was gained.

The combination of skip-gram and RNN models showed how a system could blend personalization, recommendation, and information retrieval. Personalization of recommendations via the RNN was achieved by displaying suggestions based on the course history of the student and her major. The student could add explicit personalization to this model by changing the specification of their major, deciding to turn off the collaborative based sorting of courses, or choosing to filter the sorted courses by department, degree satisfaction, registrar recommended, or by courses with open seats. Any number of these options could be selected at once. Using the subject interest and disinterest feature, a student could express a subject preference. We showed an example of this feature, accommodated by the skip-gram created vector space of courses, where courses are filtered by Economics and sorted by courses scoring high in Public Policy but not Statistics. Manual rating of the relevance of each of the nearly 4,000 courses to each of the 197 subject areas is an intractable task, but the representations of courses learned from enrollments using vector space models come with built-in relational information between courses and, by extension, between subjects by way of the centroid of their courses. This expression of preference could be framed as a query, with the results a product of information retrieval, blended with a user model when both options are turned on.

Big data presents an opportunity to venture outside of a single silo and expand the impact and scope of an adaptive system and the analyses which drive it. Access to the big data of an entire campus' historic enrollments provided this opportunity to integrate these nascent techniques rapidly into a first of its kind system that serves students from all corners of the University. We presented design considerations for recommendation systems in this domain by way of a usability study, where we found an emphasis on the importance students place on satisfying degree requirements. Feature requests were surfaced which ranged from simple in nature to implement, such as linking to descriptions of the recommended courses, to features requiring methodological research, such as explaining why a recommendation was being made. Overall, the system scored moderately high in user acceptance, with a 4.25 out of 5 average agreement when asked if they would use the service if it were available in the future. A final domain contribution was the visualization of the RNN hidden state for students in the final year of their undergraduate degree. This provided a never-before seen qualitative summary of enrollment pathways and an abstract global representation of the curricular similarities between majors.

\section{Future work}
Within the context of the recommender system, we plan to expand the evaluation in several dimensions. First, to conduct a larger user study and gauge students' impressions of the most recent functional additions. Next would be to conduct usability studies at other universities and at the community college level, where the need for scalable guidance is highest, to investigate what adaptation might be necessary based on different enrollment priorities and intuitions about the technology. Future work could also expand what is being measured in the evaluation. Affective benefits of the system, such as reduced anxieties around enrollment could be explored, as well as evaluating the system's impact on normative time to graduation. Finally, we could explore additional goodness metrics for recommendation, such as serendipity. 

In the area of scrutable connectionist user models, there is more to uncover. After revealing the semantics of what the user model believes about the user, the next step would be to investigate why the model has come to hold this belief. If the user disagrees with the model or its epistemology, she may want to change it. Therefore, enabling open user modeling paradigms within neural network frameworks is an area of future work. Also related to open user models and adaptive personalization is how to incorporate explicit goal setting by the user while still leveraging the benefits of collaborative bias.
\section*{Acknowledgements}
We would like to thank the generous contributions by UC Berkeley administrators and staff; Andrew Eppig, Mark Chiang, Max Michel, Jen Stringer, and Walter Wong with a special thanks to associate registrar Johanna Metzgar for her partnership in the deployment of the system. We would like to also thank the following undergraduate student research assistants for their contributions to the system's development; Christopher Vu Le, Andrew Joo Hun Nam, Arshad Ali Abdul Samad, Alessandra Silviera, Divyansh Agarwal, and Yuetian Luo. This work was supported by grants from the National Science Foundation (Awards 1547055 \& 1446641).

%
%

\bibliographystyle{spmpsci}      
\bibliography{bibliography}   

\begin{thebibliography}{106}
\providecommand{\natexlab}[1]{#1}
\providecommand{\url}[1]{\texttt{#1}}
\expandafter\ifx\csname urlstyle\endcsname\relax
  \providecommand{\doi}[1]{doi: #1}\else
  \providecommand{\doi}{doi: \begingroup \urlstyle{rm}\Url}\fi

\bibitem[cca(2014)]{cca}
Four-year myth.
\newblock \emph{Complete College America. Indianapolis, IN}, 2014.

\bibitem[Abdollahpouri et~al.(2017)Abdollahpouri, Burke, and
  Mobasher]{abdollahpouri2017recommender}
Himan Abdollahpouri, Robin Burke, and Bamshad Mobasher.
\newblock Recommender systems as multistakeholder environments.
\newblock In \emph{Proceedings of the 25th Conference on User Modeling,
  Adaptation and Personalization}, pages 347--348. ACM, 2017.

\bibitem[Abel et~al.(2011)Abel, Gao, Houben, and Tao]{abel2011analyzing}
Fabian Abel, Qi~Gao, Geert-Jan Houben, and Ke~Tao.
\newblock Analyzing user modeling on twitter for personalized news
  recommendations.
\newblock In Joseph Konstan, editor, \emph{User Modeling, Adaption and
  Personalization}, pages 1--12. Springer, 2011.

\bibitem[Aleven et~al.(2016)Aleven, McLaughlin, Glenn, and
  Koedinger]{aleven2016instruction}
Vincent Aleven, Elizabeth~A McLaughlin, R~Amos Glenn, and Kenneth~R Koedinger.
\newblock Instruction based on adaptive learning technologies.
\newblock In R.~E. Mayer and P.~Alexander, editors, \emph{Handbook of research
  on learning and instruction. Routledge}, 2016.

\bibitem[Arnold and Pistilli(2012)]{arnold2012course}
Kimberly~E Arnold and Matthew~D Pistilli.
\newblock Course signals at purdue: Using learning analytics to increase
  student success.
\newblock In \emph{Proceedings of the 2nd international conference on learning
  analytics and knowledge}, pages 267--270. ACM, 2012.

\bibitem[Barkan and Koenigstein(2016)]{barkan2016item2vec}
Oren Barkan and Noam Koenigstein.
\newblock Item2vec: neural item embedding for collaborative filtering.
\newblock In \emph{Machine Learning for Signal Processing (MLSP), 2016 IEEE
  26th International Workshop on}, pages 1--6. IEEE, 2016.

\bibitem[Bastien et~al.(2012)Bastien, Lamblin, Pascanu, Bergstra, Goodfellow,
  Bergeron, Bouchard, Warde-Farley, and Bengio]{bastien2012theano}
Fr{\'e}d{\'e}ric Bastien, Pascal Lamblin, Razvan Pascanu, James Bergstra, Ian
  Goodfellow, Arnaud Bergeron, Nicolas Bouchard, David Warde-Farley, and Yoshua
  Bengio.
\newblock Theano: new features and speed improvements.
\newblock \emph{arXiv preprint arXiv:1211.5590}, 2012.

\bibitem[Bayer et~al.(2017)Bayer, He, Kanagal, and Rendle]{bayer2017generic}
Immanuel Bayer, Xiangnan He, Bhargav Kanagal, and Steffen Rendle.
\newblock A generic coordinate descent framework for learning from implicit
  feedback.
\newblock In \emph{Proceedings of the 26th International Conference on World
  Wide Web}, pages 1341--1350. International World Wide Web Conferences
  Steering Committee, 2017.

\bibitem[Bengio et~al.(2003)Bengio, Ducharme, Vincent, and
  Jauvin]{bengio2003neural}
Yoshua Bengio, R{\'e}jean Ducharme, Pascal Vincent, and Christian Jauvin.
\newblock A neural probabilistic language model.
\newblock In Jaz Kandola, Thomas Hofmann, Tomaso Poggio, and John Shawe-Taylor,
  editors, \emph{Journal of machine learning research}, volume~3, pages
  1137--1155, 2003.

\bibitem[Bergstra et~al.(2010)Bergstra, Breuleux, Bastien, Lamblin, Pascanu,
  Desjardins, Turian, Warde-Farley, and Bengio]{bergstra2010theano}
James Bergstra, Olivier Breuleux, Fr{\'e}d{\'e}ric Bastien, Pascal Lamblin,
  Razvan Pascanu, Guillaume Desjardins, Joseph Turian, David Warde-Farley, and
  Yoshua Bengio.
\newblock Theano: A cpu and gpu math compiler in python.
\newblock In \emph{Proc. 9th Python in Science Conf}, pages 1--7, 2010.

\bibitem[Berkovsky et~al.(2007)Berkovsky, Kuflik, and
  Ricci]{berkovsky2007cross}
Shlomo Berkovsky, Tsvi Kuflik, and Francesco Ricci.
\newblock Cross-domain mediation in collaborative filtering.
\newblock In \emph{International Conference on User Modeling}, pages 355--359.
  Springer, 2007.

\bibitem[Bolukbasi et~al.(2016)Bolukbasi, Chang, Zou, Saligrama, and
  Kalai]{bolukbasi2016man}
Tolga Bolukbasi, Kai-Wei Chang, James~Y Zou, Venkatesh Saligrama, and Adam~T
  Kalai.
\newblock Man is to computer programmer as woman is to homemaker? debiasing
  word embeddings.
\newblock In \emph{Advances in Neural Information Processing Systems}, pages
  4349--4357, 2016.

\bibitem[Brown et~al.(1992)Brown, Desouza, Mercer, Pietra, and
  Lai]{brown1992class}
Peter~F Brown, Peter~V Desouza, Robert~L Mercer, Vincent J~Della Pietra, and
  Jenifer~C Lai.
\newblock Class-based n-gram models of natural language.
\newblock \emph{Computational linguistics}, 18\penalty0 (4):\penalty0 467--479,
  1992.

\bibitem[Brusilovsky(1996)]{brusilovsky1996methods}
Peter Brusilovsky.
\newblock Methods and techniques of adaptive hypermedia.
\newblock \emph{User modeling and user-adapted interaction}, 6\penalty0
  (2-3):\penalty0 87--129, 1996.

\bibitem[Bull and Kay(2010)]{bull2010open}
Susan Bull and Judy Kay.
\newblock Open learner models.
\newblock In R.~Nkambou, editor, \emph{Advances in intelligent tutoring
  systems}, pages 301--322. Springer, 2010.

\bibitem[{Burke}(2002)]{burke2002hybrid}
Robin~D. {Burke}.
\newblock Hybrid recommender systems: Survey and experiments.
\newblock In Ingrid Zukerman, editor, \emph{User Modeling and User-adapted
  Interaction}, volume~12, pages 331--370, 2002.

\bibitem[Burrell(2016)]{burrell2016machine}
Jenna Burrell.
\newblock How the machine ‘thinks’: Understanding opacity in machine
  learning algorithms.
\newblock \emph{Big Data \& Society}, 3\penalty0 (1):\penalty0
  2053951715622512, 2016.

\bibitem[Carmagnola et~al.()Carmagnola, Vernero, and
  Grillo]{carmagnola2009sonars}
Francesca Carmagnola, Fabiana Vernero, and Pierluigi Grillo.
\newblock Sonars: A social networks-based algorithm for social recommender
  systems.
\newblock In G.-J.~Houben et~al., editor, \emph{User Modeling, Adaptation, and
  Personalization}.

\bibitem[Chaturapruek et~al.(2018)Chaturapruek, Dee, Johari, Kizilcec, and
  Stevens]{chaturapruek2018data}
Sorathan Chaturapruek, Thomas Dee, Ramesh Johari, Ren{\'e} Kizilcec, and
  Mitchell Stevens.
\newblock How a data-driven course planning tool affects college students' gpa:
  evidence from two field experiments.
\newblock 2018.

\bibitem[Chen(2018)]{chen2018behavior2vec}
Hung-Hsuan Chen.
\newblock Behavior2vec: Generating distributed representations of users’
  behaviors on products for recommender systems.
\newblock \emph{ACM Transactions on Knowledge Discovery from Data (TKDD)},
  12\penalty0 (4):\penalty0 43, 2018.

\bibitem[Chollet et~al.(2015)]{chollet2015keras}
Fran\c{c}ois Chollet et~al.
\newblock Keras.
\newblock \url{https://keras.io}, 2015.

\bibitem[Corbett(2001)]{corbett2001cognitive}
Albert Corbett.
\newblock Cognitive computer tutors: Solving the two-sigma problem.
\newblock \emph{International Conference on User Modeling}, pages 137--147,
  2001.

\bibitem[Cragun and Day(1999)]{cragun1999dynamic}
Brian~John Cragun and Paul~Reuben Day.
\newblock Dynamic regulation of television viewing content based on viewer
  profile and viewing history, October~26 1999.
\newblock US Patent 5,973,683.

\bibitem[Czarkowski and Kay(2000)]{czarkowski2000bringing}
Marek Czarkowski and Judy Kay.
\newblock Bringing scrutability to adaptive hypertext teaching.
\newblock In \emph{International Conference on Intelligent Tutoring Systems},
  pages 423--432. Springer, 2000.

\bibitem[Czarkowski et~al.(2005)Czarkowski, Kay, and
  Potts]{czarkowski2005scrutability}
Marek Czarkowski, Judy Kay, and Serena Potts.
\newblock Scrutability as a core interface element.
\newblock In \emph{Proceedings of the 2005 conference on Artificial
  Intelligence in Education: Supporting Learning through Intelligent and
  Socially Informed Technology}, pages 783--785. IOS Press, 2005.

\bibitem[DeAngelo et~al.(2011)DeAngelo, Franke, Hurtado, Pryor, and
  Tran]{deangelo2011completing}
Linda DeAngelo, Ray Franke, Sylvia Hurtado, John~H Pryor, and Serge Tran.
\newblock Completing college: Assessing graduation rates at four-year
  institutions.
\newblock \emph{Los Angeles: Higher Education Research Institute, UCLA}, 2011.

\bibitem[Devooght and Bersini(2017)]{devooght2017long}
Robin Devooght and Hugues Bersini.
\newblock Long and short-term recommendations with recurrent neural networks.
\newblock In \emph{Proceedings of the 25th Conference on User Modeling,
  Adaptation and Personalization}, pages 13--21. ACM, 2017.

\bibitem[{Diamond} and {Lee}(2011)]{diamond2011interventions}
Adele {Diamond} and Kathleen {Lee}.
\newblock Interventions shown to aid executive function development in children
  4 to 12 years old.
\newblock \emph{Science}, 333\penalty0 (6045):\penalty0 959--964, 2011.

\bibitem[Elbadrawy and Karypis(2016)]{elbadrawy2016domain}
Asmaa Elbadrawy and George Karypis.
\newblock Domain-aware grade prediction and top-n course recommendation.
\newblock In \emph{RecSys}, pages 183--190, 2016.

\bibitem[Fan et~al.(2014)Fan, Qian, Xie, and Soong]{fan2014tts}
Yuchen Fan, Yao Qian, Feng-Long Xie, and Frank~K Soong.
\newblock Tts synthesis with bidirectional lstm based recurrent neural
  networks.
\newblock In \emph{Fifteenth Annual Conference of the International Speech
  Communication Association}, 2014.

\bibitem[Farzan and Brusilovsky(2011)]{farzan2011encouraging}
Rosta Farzan and Peter Brusilovsky.
\newblock Encouraging user participation in a course recommender system: An
  impact on user behavior.
\newblock \emph{Computers in Human Behavior}, 27\penalty0 (1):\penalty0
  276--284, 2011.

\bibitem[Gers et~al.(1999)Gers, Schmidhuber, and Cummins]{gers1999learning}
Felix~A Gers, J{\"u}rgen Schmidhuber, and Fred Cummins.
\newblock Learning to forget: Continual prediction with lstm.
\newblock In \emph{9th International Conference on Artificial Neural Networks},
  pages 850--855, 1999.

\bibitem[Goldberg(2016)]{goldberg2016primer}
Yoav Goldberg.
\newblock A primer on neural network models for natural language processing.
\newblock \emph{Journal of Artificial Intelligence Research}, 57:\penalty0
  345--420, 2016.

\bibitem[Graves et~al.(2013)Graves, Mohamed, and Hinton]{graves2013speech}
Alex Graves, Abdel-rahman Mohamed, and Geoffrey Hinton.
\newblock Speech recognition with deep recurrent neural networks.
\newblock In \emph{International Conference on Acoustics, Speech and Signal
  Processing}, pages 6645--6649. IEEE, 2013.

\bibitem[Grbovic et~al.(2015)Grbovic, Radosavljevic, Djuric, Bhamidipati,
  Savla, Bhagwan, and Sharp]{grbovic2015commerce}
Mihajlo Grbovic, Vladan Radosavljevic, Nemanja Djuric, Narayan Bhamidipati,
  Jaikit Savla, Varun Bhagwan, and Doug Sharp.
\newblock E-commerce in your inbox: Product recommendations at scale.
\newblock In \emph{Proceedings of the 21th International Conference on
  Knowledge Discovery and Data Mining}, pages 1809--1818. ACM, 2015.

\bibitem[Gregor et~al.(2015)Gregor, Danihelka, Graves, Rezende, and
  Wierstra]{gregor2015draw}
Karol Gregor, Ivo Danihelka, Alex Graves, Danilo~Jimenez Rezende, and Daan
  Wierstra.
\newblock Draw: A recurrent neural network for image generation.
\newblock \emph{arXiv preprint arXiv:1502.04623}, 2015.

\bibitem[Gu{\`a}rdia-Sebaoun et~al.(2015)Gu{\`a}rdia-Sebaoun, Guigue, and
  Gallinari]{guardia2015latent}
Elie Gu{\`a}rdia-Sebaoun, Vincent Guigue, and Patrick Gallinari.
\newblock Latent trajectory modeling: A light and efficient way to introduce
  time in recommender systems.
\newblock In \emph{Proceedings of the 9th ACM Conference on Recommender
  Systems}, pages 281--284. ACM, 2015.

\bibitem[Guo et~al.(2012)Guo, Zhang, and Thalmann]{guo2012simple}
Guibing Guo, Jie Zhang, and Daniel Thalmann.
\newblock A simple but effective method to incorporate trusted neighbors in
  recommender systems.
\newblock In \emph{International Conference on User Modeling, Adaptation, and
  Personalization}, pages 114--125. Springer, 2012.

\bibitem[He et~al.(2016)He, Zhang, Kan, and Chua]{he2016fast}
Xiangnan He, Hanwang Zhang, Min-Yen Kan, and Tat-Seng Chua.
\newblock Fast matrix factorization for online recommendation with implicit
  feedback.
\newblock In \emph{Proceedings of the 39th International Conference on Research
  and Development in Information Retrieval}, pages 549--558. ACM, 2016.

\bibitem[He et~al.(2017)He, Liao, Zhang, Nie, Hu, and Chua]{he2017neural}
Xiangnan He, Lizi Liao, Hanwang Zhang, Liqiang Nie, Xia Hu, and Tat-Seng Chua.
\newblock Neural collaborative filtering.
\newblock In \emph{Proceedings of the 26th International Conference on World
  Wide Web}, pages 173--182. International World Wide Web Conferences Steering
  Committee, 2017.

\bibitem[Hidasi et~al.(2016)Hidasi, Karatzoglou, Baltrunas, and
  Tikk]{hidasi2015session}
Bal{\'a}zs Hidasi, Alexandros Karatzoglou, Linas Baltrunas, and Domonkos Tikk.
\newblock Session-based recommendations with recurrent neural networks.
\newblock \emph{arXiv preprint arXiv:1511.06939, ICLR}, 2016.

\bibitem[Hinton(1986)]{hinton1986learning}
Geoffrey~E Hinton.
\newblock Learning distributed representations of concepts.
\newblock In \emph{Proceedings of the 8-th Annual Conference of the Cognitive
  Science Society}, volume~1, page~12. Amherst, MA, 1986.

\bibitem[Hochreiter and Schmidhuber(1997)]{hochreiter1997long}
Sepp Hochreiter and J{\"u}rgen Schmidhuber.
\newblock Long short-term memory.
\newblock \emph{Neural Computation}, 9\penalty0 (8):\penalty0 1735--1780, 1997.

\bibitem[Hodara et~al.(2016)Hodara, Martinez-Wenzl, Stevens, and
  Mazzeo]{hodara2016improving}
Michelle Hodara, Mary Martinez-Wenzl, David Stevens, and Christopher Mazzeo.
\newblock Improving credit mobility for community college transfer students.
\newblock In Sidney Hacker, Kelly DeForrest, Allyson Hagen, Rhonda Barton, and
  Glenn Wright, editors, \emph{Planning for Higher Education}, volume~45,
  page~51. Society for College and University Planning, 2016.

\bibitem[Jannach et~al.(2017)Jannach, Ludewig, and Lerche]{jannach2017session}
Dietmar Jannach, Malte Ludewig, and Lukas Lerche.
\newblock Session-based item recommendation in e-commerce: on short-term
  intents, reminders, trends and discounts.
\newblock \emph{User Modeling and User-Adapted Interaction}, pages 1--42, 2017.

\bibitem[Jayaprakash et~al.(2014)Jayaprakash, Moody, Laur{\'\i}a, Regan, and
  Baron]{jayaprakash2014early}
Sandeep~M Jayaprakash, Erik~W Moody, Eitel~JM Laur{\'\i}a, James~R Regan, and
  Joshua~D Baron.
\newblock Early alert of academically at-risk students: An open source
  analytics initiative.
\newblock \emph{Journal of Learning Analytics}, 1\penalty0 (1):\penalty0 6--47,
  2014.

\bibitem[Kay(2000)]{kay2000stereotypes}
Judy Kay.
\newblock Stereotypes, student models and scrutability.
\newblock In \emph{Intelligent Tutoring Systems}, volume 1839, pages 19--30.
  Springer, 2000.

\bibitem[Kay and Kummerfeld(2006)]{kay2006scrutability}
Judy Kay and Bob Kummerfeld.
\newblock Scrutability, user control and privacy for distributed
  personalization.
\newblock In \emph{Proceedings of the CHI Workshop on Privacy-Enhanced
  Personalization}, pages 21--22, 2006.

\bibitem[Kingma and Ba(2014)]{kingma2014adam}
Diederik Kingma and Jimmy Ba.
\newblock Adam: A method for stochastic optimization.
\newblock \emph{arXiv preprint arXiv:1412.6980}, 2014.

\bibitem[Konstan and Riedl(2012)]{konstan2012recommender}
Joseph~A Konstan and John Riedl.
\newblock Recommender systems: from algorithms to user experience.
\newblock \emph{User Modeling and User-Adapted Interaction}, 22\penalty0
  (1):\penalty0 101--123, 2012.

\bibitem[Koren(2010{\natexlab{a}})]{koren2010collaborative}
Yehuda Koren.
\newblock Collaborative filtering with temporal dynamics.
\newblock In \emph{Proceedings of the 15th ACM SIGKDD International Conference
  on Knowledge Discovery and Data Mining}, volume~53, pages 89--97. ACM,
  2010{\natexlab{a}}.

\bibitem[Koren(2010{\natexlab{b}})]{koren2010factor}
Yehuda Koren.
\newblock Factor in the neighbors: Scalable and accurate collaborative
  filtering.
\newblock \emph{ACM Transactions on Knowledge Discovery from Data (TKDD)},
  4\penalty0 (1):\penalty0 1, 2010{\natexlab{b}}.

\bibitem[Koren and Bell(2015)]{koren2015advances}
Yehuda Koren and Robert Bell.
\newblock Advances in collaborative filtering.
\newblock In \emph{Recommender Systems Handbook}, pages 77--118. Springer,
  2015.

\bibitem[Koren et~al.(2009)Koren, Bell, and Volinsky]{koren2009matrix}
Yehuda Koren, Robert Bell, and Chris Volinsky.
\newblock Matrix factorization techniques for recommender systems.
\newblock \emph{Computer}, 42\penalty0 (8), 2009.

\bibitem[Kouki et~al.(2017)Kouki, Schaffer, Pujara, O'Donovan, and
  Getoor]{KoukiSPOG17}
Pigi Kouki, James Schaffer, Jay Pujara, John O'Donovan, and Lise Getoor.
\newblock User preferences for hybrid explanations.
\newblock In \emph{Proceedings of the Eleventh ACM Conference on Recommender
  Systems}, pages 84--88. ACM, 2017.

\bibitem[Krizhevsky et~al.(2012)Krizhevsky, Sutskever, and
  Hinton]{krizhevsky2012imagenet}
Alex Krizhevsky, Ilya Sutskever, and Geoffrey~E Hinton.
\newblock Imagenet classification with deep convolutional neural networks.
\newblock In \emph{Advances in neural information processing systems}, pages
  1097--1105, 2012.

\bibitem[Kuusela and Pallab(2000)]{kuusela2000comparison}
Hannu Kuusela and Paul Pallab.
\newblock A comparison of concurrent and retrospective verbal protocol
  analysis.
\newblock \emph{The American Journal of Psychology}, 113\penalty0 (3):\penalty0
  387, 2000.

\bibitem[Kyriacou et~al.(2009)Kyriacou, Davis, and
  Tiropanis]{kyriacou2009evaluating}
Demetris Kyriacou, Hugh~C Davis, and Thanassis Tiropanis.
\newblock Evaluating three scrutability and three privacy user privileges for a
  scrutable user modelling infrastructure.
\newblock In \emph{International Conference on User Modeling, Adaptation, and
  Personalization}, pages 428--434. Springer, 2009.

\bibitem[Lai et~al.(2015)Lai, Xu, Liu, and Zhao]{lai2015recurrent}
Siwei Lai, Liheng Xu, Kang Liu, and Jun Zhao.
\newblock Recurrent convolutional neural networks for text classification.
\newblock In \emph{AAAI}, volume 333, pages 2267--2273, 2015.

\bibitem[LeCun et~al.(2015)LeCun, Bengio, and Hinton]{lecun2015deep}
Yann LeCun, Yoshua Bengio, and Geoffrey Hinton.
\newblock Deep learning.
\newblock \emph{Nature}, 521\penalty0 (7553):\penalty0 436--444, 2015.

\bibitem[Levy and Goldberg(2014{\natexlab{a}})]{levy2014dependency}
Omer Levy and Yoav Goldberg.
\newblock Dependency-based word embeddings.
\newblock In \emph{Proceedings of the 52nd Annual Meeting of the Association
  for Computational Linguistics}, pages 302--308, 2014{\natexlab{a}}.

\bibitem[Levy and Goldberg(2014{\natexlab{b}})]{levy2014neural}
Omer Levy and Yoav Goldberg.
\newblock Neural word embedding as implicit matrix factorization.
\newblock In \emph{Advances in Neural Information Processing Systems}, pages
  2177--2185, 2014{\natexlab{b}}.

\bibitem[Li et~al.(2012)Li, Tinapple, and Sundaram]{li2012visual}
Zhen Li, David Tinapple, and Hari Sundaram.
\newblock Visual planner: beyond prerequisites, designing an interactive course
  planner for a 21st century flexible curriculum.
\newblock In \emph{CHI'12 Extended Abstracts on Human Factors in Computing
  Systems}, pages 1613--1618. ACM, 2012.

\bibitem[Liang et~al.(2016)Liang, Charlin, McInerney, and
  Blei]{liang2016modeling}
Dawen Liang, Laurent Charlin, James McInerney, and David~M Blei.
\newblock Modeling user exposure in recommendation.
\newblock In \emph{Proceedings of the 25th International Conference on World
  Wide Web}, pages 951--961. International World Wide Web Conferences Steering
  Committee, 2016.

\bibitem[Linden et~al.(2003)Linden, Smith, and York]{linden2003amazon}
Greg Linden, Brent Smith, and Jeremy York.
\newblock Amazon. com recommendations: Item-to-item collaborative filtering.
\newblock \emph{IEEE Internet computing}, 7\penalty0 (1):\penalty0 76--80,
  2003.

\bibitem[Luo and Pardos(2018)]{luo2018diag}
Yutian Luo and Zachary~Alexander Pardos.
\newblock Diagnosing university student subject proficiency and predicting
  degree completion in vector space.
\newblock In E.~Eaton and M.~Wollowski, editors, \emph{AAAI}, pages 7920--7927,
  2018.

\bibitem[Mairal et~al.(2010)Mairal, Bach, Ponce, and Sapiro]{mairal2010online}
Julien Mairal, Francis Bach, Jean Ponce, and Guillermo Sapiro.
\newblock Online learning for matrix factorization and sparse coding.
\newblock In Hui Zou, editor, \emph{Journal of Machine Learning Research},
  volume~11, pages 19--60, 2010.

\bibitem[Mao et~al.(2014)Mao, Xu, Yang, Wang, Huang, and Yuille]{mao2014deep}
Junhua Mao, Wei Xu, Yi~Yang, Jiang Wang, Zhiheng Huang, and Alan Yuille.
\newblock Deep captioning with multimodal recurrent neural networks (m-rnn).
\newblock \emph{arXiv preprint arXiv:1412.6632}, 2014.

\bibitem[Mikolov et~al.(2010)Mikolov, Karafi{\'a}t, Burget, Cernock{\`y}, and
  Khudanpur]{mikolov2010recurrent}
Tomas Mikolov, Martin Karafi{\'a}t, Lukas Burget, Jan Cernock{\`y}, and Sanjeev
  Khudanpur.
\newblock Recurrent neural network based language model.
\newblock In \emph{11th Annual Conference of the International Speech
  Communication Association}, volume~2, page~3, 2010.

\bibitem[Mikolov et~al.(2013{\natexlab{a}})Mikolov, Chen, Corrado, and
  Dean]{mikolov2013efficient}
Tomas Mikolov, Kai Chen, Greg Corrado, and Jeffrey Dean.
\newblock Efficient estimation of word representations in vector space.
\newblock \emph{arXiv preprint arXiv:1301.3781}, 2013{\natexlab{a}}.

\bibitem[Mikolov et~al.(2013{\natexlab{b}})Mikolov, Sutskever, Chen, Corrado,
  and Dean]{mikolov2013distributed}
Tomas Mikolov, Ilya Sutskever, Kai Chen, Greg~S Corrado, and Jeff Dean.
\newblock Distributed representations of words and phrases and their
  compositionality.
\newblock In \emph{Advances in neural information processing systems}, pages
  3111--3119, 2013{\natexlab{b}}.

\bibitem[Mnih et~al.(2015)Mnih, Kavukcuoglu, Silver, Rusu, Veness, Bellemare,
  Graves, Riedmiller, Fidjeland, Ostrovski, et~al.]{mnih2015human}
Volodymyr Mnih, Koray Kavukcuoglu, David Silver, Andrei~A Rusu, Joel Veness,
  Marc~G Bellemare, Alex Graves, Martin Riedmiller, Andreas~K Fidjeland, Georg
  Ostrovski, et~al.
\newblock Human-level control through deep reinforcement learning.
\newblock \emph{Nature}, 518\penalty0 (7540):\penalty0 529--533, 2015.

\bibitem[Musto et~al.(2016)Musto, Semeraro, de~Gemmis, and
  Lops]{musto2016learning}
Cataldo Musto, Giovanni Semeraro, Marco de~Gemmis, and Pasquale Lops.
\newblock Learning word embeddings from wikipedia for content-based recommender
  systems.
\newblock In \emph{European Conference on Information Retrieval}, pages
  729--734. Springer, 2016.

\bibitem[Parameswaran et~al.(2011)Parameswaran, Venetis, and
  Garcia-Molina]{parameswaran2011recommendation}
Aditya Parameswaran, Petros Venetis, and Hector Garcia-Molina.
\newblock Recommendation systems with complex constraints: A course
  recommendation perspective.
\newblock \emph{ACM Transactions on Information Systems (TOIS)}, 29\penalty0
  (4):\penalty0 20, 2011.

\bibitem[Pardos and Dadu(2017)]{pardos2017imputing}
Zachary~A Pardos and Anant Dadu.
\newblock Imputing {KC}s with representations of problem content and context.
\newblock In Federica Cena and Michel Desmarias, editors, \emph{Proceedings of
  the 25th Conference on User Modeling, Adaptation and Personalization}, pages
  148--155. ACM, 2017.

\bibitem[Pardos and Horodyskyj(2017)]{pardos2017analysis}
Zachary~A Pardos and Lev Horodyskyj.
\newblock Analysis of student behaviour in habitable worlds using continuous
  representation visualization.
\newblock \emph{CoRR preprint, abs/1710.06654}, 2017.
\newblock URL \url{https://arxiv.org/abs/1710.06654}.

\bibitem[Pardos and Nam(2017)]{pardos2017school}
Zachary~A Pardos and Andrew Joo~Hun Nam.
\newblock The school of information and its relationship to computer science at
  uc berkeley.
\newblock \emph{iConference 2017 Proceedings}, 2017.

\bibitem[Pardos and Nam(2018)]{pardos2018map}
Zachary~A Pardos and Andrew Joo~Hun Nam.
\newblock A map of knowledge.
\newblock \emph{CoRR preprint, abs/1811.07974}, 2018.
\newblock URL \url{https://arxiv.org/abs/1811.07974}.

\bibitem[Pardos et~al.(2017)Pardos, Tang, Davis, and Le]{pardos2017enabling}
Zachary~A Pardos, Steven Tang, Daniel Davis, and Christopher~Vu Le.
\newblock Enabling real-time adaptivity in moocs with a personalized next-step
  recommendation framework.
\newblock In C.~Thille and J.~Reich, editors, \emph{Proceedings of the Fourth
  (2017) ACM Conference on Learning@ Scale}, pages 23--32. ACM, 2017.

\bibitem[Pardos et~al.(2018)Pardos, Farrar, Academy, Kolb, Peh, and
  Lee]{Pardos2018}
Zachary~A. Pardos, Scott Farrar, Khan Academy, John Kolb, Gao~Xian Peh, and
  Jong~Ha Lee.
\newblock {Distributed Representation of Misconceptions}.
\newblock In \emph{Proceedings of the 13th International Conference of the
  Learning Sciences (ICLS)}, pages 1791--1798, London, UK, 2018.

\bibitem[Paterek(2007)]{paterek2007improving}
Arkadiusz Paterek.
\newblock Improving regularized singular value decomposition for collaborative
  filtering.
\newblock In \emph{Proceedings of KDD cup and workshop}, volume 2007, pages
  5--8, 2007.

\bibitem[Pazzani(1999)]{pazzani1999framework}
Michael~J Pazzani.
\newblock A framework for collaborative, content-based and demographic
  filtering.
\newblock \emph{Artificial Intelligence Review}, 13\penalty0 (5-6):\penalty0
  393--408, 1999.

\bibitem[Pazzani and Billsus(2007)]{pazzani2007content}
Michael~J Pazzani and Daniel Billsus.
\newblock Content-based recommendation systems.
\newblock In \emph{The adaptive web}, pages 325--341. Springer, 2007.

\bibitem[Phelan et~al.(2009)Phelan, McCarthy, and Smyth]{phelan2009using}
Owen Phelan, Kevin McCarthy, and Barry Smyth.
\newblock Using twitter to recommend real-time topical news.
\newblock In \emph{Proceedings of the third ACM conference on Recommender
  systems}, pages 385--388. ACM, 2009.

\bibitem[Pu et~al.(2011)Pu, Chen, and Hu]{pu2011user}
Pearl Pu, Li~Chen, and Rong Hu.
\newblock A user-centric evaluation framework for recommender systems.
\newblock In \emph{Proceedings of the fifth ACM conference on Recommender
  systems}, pages 157--164. ACM, 2011.

\bibitem[Rashid et~al.(2002)Rashid, Albert, Cosley, Lam, McNee, Konstan, and
  Riedl]{rashid2002getting}
Al~Mamunur Rashid, Istvan Albert, Dan Cosley, Shyong~K Lam, Sean~M McNee,
  Joseph~A Konstan, and John Riedl.
\newblock Getting to know you: learning new user preferences in recommender
  systems.
\newblock In \emph{Proceedings of the 7th International Conference on
  Intelligent User Interfaces}, pages 127--134. ACM, 2002.

\bibitem[{\v R}eh{\r u}{\v r}ek and Sojka(2010)]{rehurek_lrec}
Radim {\v R}eh{\r u}{\v r}ek and Petr Sojka.
\newblock {Software Framework for Topic Modelling with Large Corpora}.
\newblock In \emph{{Proceedings of the LREC 2010 Workshop on New Challenges for
  NLP Frameworks}}, pages 45--50, Valletta, Malta, May 2010. ELRA.
\newblock \url{http://is.muni.cz/publication/884893/en}.

\bibitem[Sacin et~al.()Sacin, Agapito, Shafti, and
  Ortigosa]{sacin2009recommendation}
Cesar~Vialardi Sacin, Javier~Bravo Agapito, Leila Shafti, and Alvaro Ortigosa.
\newblock Recommendation in higher education using data mining techniques.
\newblock In \emph{International Working Group on Educational Data Mining}.

\bibitem[Schafer et~al.(2007)Schafer, Frankowski, Herlocker, and
  Sen]{schafer2007collaborative}
J~Ben Schafer, Dan Frankowski, Jon Herlocker, and Shilad Sen.
\newblock Collaborative filtering recommender systems.
\newblock In \emph{The Adaptive Web}, pages 291--324. Springer, 2007.

\bibitem[Shapiro et~al.(2017)Shapiro, Dundar, Huie, Wakhungu, Yuan, Nathan, and
  Bhimdiwala]{shapiro2017completing}
Doug Shapiro, Afet Dundar, Faye Huie, Phoebe~Khasiala Wakhungu, Xin Yuan, Angel
  Nathan, and Ayesha Bhimdiwala.
\newblock Completing college: A national view of student completion rates--fall
  2011 cohort.(signature report no. 14).
\newblock \emph{National Student Clearinghouse}, 2017.

\bibitem[Siemens et~al.(2011)Siemens, Gasevic, Haythornthwaite, Dawson, Shum,
  Ferguson, Duval, Verbert, and Baker]{siemens2011open}
George Siemens, Dragan Gasevic, Caroline Haythornthwaite, Shane Dawson,
  S~Buckingham Shum, Rebecca Ferguson, Erik Duval, Katrien Verbert, and RSJD
  Baker.
\newblock \emph{Open Learning Analytics: an integrated \& modularized
  platform}.
\newblock PhD thesis, Open University Press Doctoral dissertation, 2011.

\bibitem[Snow et~al.(2015)Snow, Allen, Jacovina, and McNamara]{snow2015does}
Erica~L Snow, Laura~K Allen, Matthew~E Jacovina, and Danielle~S McNamara.
\newblock Does agency matter?: Exploring the impact of controlled behaviors
  within a game-based environment.
\newblock \emph{Computers \& Education}, 82:\penalty0 378--392, 2015.

\bibitem[Suglia et~al.(2017)Suglia, Greco, Musto, de~Gemmis, Lops, and
  Semeraro]{suglia2017deep}
Alessandro Suglia, Claudio Greco, Cataldo Musto, Marco de~Gemmis, Pasquale
  Lops, and Giovanni Semeraro.
\newblock A deep architecture for content-based recommendations exploiting
  recurrent neural networks.
\newblock In \emph{Proceedings of the 25th Conference on User Modeling,
  Adaptation and Personalization}, pages 202--211. ACM, 2017.

\bibitem[Sutskever et~al.(2014)Sutskever, Vinyals, and
  Le]{sutskever2014sequence}
Ilya Sutskever, Oriol Vinyals, and Quoc~V Le.
\newblock Sequence to sequence learning with neural networks.
\newblock In \emph{Advances in neural information processing systems}, pages
  3104--3112, 2014.

\bibitem[Tan et~al.(2016)Tan, Xu, and Liu]{tan2016improved}
Yong~Kiam Tan, Xinxing Xu, and Yong Liu.
\newblock Improved recurrent neural networks for session-based recommendations.
\newblock In \emph{Proceedings of the 1st Workshop on Deep Learning for
  Recommender Systems}, pages 17--22. ACM, 2016.

\bibitem[Tang et~al.(2017)Tang, Peterson, and Pardos]{tang2016modelling}
Steven Tang, Joshua~C Peterson, and Zachary~A Pardos.
\newblock Predictive modelling of student behaviour using granular large-scale
  action data.
\newblock \emph{The Handbook of Learning Analytics}, pages 223--233, 2017.

\bibitem[Van Der~Maaten()]{van2014accelerating}
Laurens Van Der~Maaten.
\newblock Accelerating t-sne using tree-based algorithms.
\newblock In Aaron Courville, Rob Fergus, and Christopher Manning, editors,
  \emph{Journal of Machine Learning Research}.

\bibitem[Van~Meteren and Van~Someren(2000)]{van2000using}
Robin Van~Meteren and Maarten Van~Someren.
\newblock Using content-based filtering for recommendation.
\newblock In \emph{Proceedings of the Machine Learning in the New Information
  Age: MLnet/ECML2000 Workshop}, pages 47--56, 2000.

\bibitem[Wang et~al.(2017)Wang, Sy, Liu, and Piech]{wang2017learning}
Lisa Wang, Angela Sy, Larry Liu, and Chris Piech.
\newblock Learning to represent student knowledge on programming exercises
  using deep learning.
\newblock In \emph{Proceedings of the 10th International Conference on
  Educational Data Mining; Wuhan, China}, pages 324--329, 2017.

\bibitem[Wang et~al.(2018)Wang, He, Feng, Nie, and Chua]{wang2018tem}
Xiang Wang, Xiangnan He, Fuli Feng, Liqiang Nie, and Tat-Seng Chua.
\newblock Tem: Tree-enhanced embedding model for explainable recommendation.
\newblock In \emph{Proceedings of the 2018 World Wide Web Conference on World
  Wide Web}, pages 1543--1552. International World Wide Web Conferences
  Steering Committee, 2018.

\bibitem[Whitehill et~al.(2017)Whitehill, Mohan, Seaton, Rosen, and
  Tingley]{whitehill2017delving}
Jacob Whitehill, Kiran Mohan, Daniel Seaton, Yigal Rosen, and Dustin Tingley.
\newblock Delving deeper into mooc student dropout prediction.
\newblock \emph{arXiv preprint arXiv:1702.06404}, 2017.

\bibitem[Xu et~al.(2015)Xu, Ba, Kiros, Cho, Courville, Salakhudinov, Zemel, and
  Bengio]{xu2015show}
Kelvin Xu, Jimmy Ba, Ryan Kiros, Kyunghyun Cho, Aaron Courville, Ruslan
  Salakhudinov, Rich Zemel, and Yoshua Bengio.
\newblock Show, attend and tell: Neural image caption generation with visual
  attention.
\newblock In \emph{International Conference on Machine Learning}, pages
  2048--2057, 2015.

\bibitem[Yang et~al.(2013)Yang, Sinha, Adamson, and Ros{\'e}]{yang2013turn}
Diyi Yang, Tanmay Sinha, David Adamson, and Carolyn~Penstein Ros{\'e}.
\newblock Turn on, tune in, drop out: Anticipating student dropouts in massive
  open online courses.
\newblock In \emph{Proceedings of the 2013 NIPS Data-driven education
  workshop}, volume~11, page~14, 2013.

\bibitem[Yu et~al.(2006)Yu, Zhou, Hao, and Gu]{yu2006tv}
Zhiwen Yu, Xingshe Zhou, Yanbin Hao, and Jianhua Gu.
\newblock Tv program recommendation for multiple viewers based on user profile
  merging.
\newblock \emph{User Modeling and User-adapted Interaction}, 16\penalty0
  (1):\penalty0 63--82, 2006.

\bibitem[Zanotti et~al.(2016)Zanotti, Horvath, Barbosa, Immedisetty, and
  Gemmell]{zanotti2016infusing}
Greg Zanotti, Miller Horvath, Lucas~Nunes Barbosa, Venkata Trinadh Kumar~Gupta
  Immedisetty, and Jonathan Gemmell.
\newblock Infusing collaborative recommenders with distributed representations.
\newblock In \emph{Proceedings of the 1st Workshop on Deep Learning for
  Recommender Systems}, pages 35--42. ACM, 2016.

\bibitem[Zimmerman(1990)]{zimmerman1990self}
Barry~J Zimmerman.
\newblock Self-regulated learning and academic achievement: An overview.
\newblock \emph{Educational psychologist}, 25\penalty0 (1):\penalty0 3--17,
  1990.

\end{thebibliography}
\clearpage

%
%

\end{document}